\documentclass{article} % For LaTeX2e
\usepackage{iclr2027_preprint,times}
\renewcommand{\headrulewidth}{0pt}

\usepackage{amsmath,amsfonts,bm}

\def\eqref#1{equation~\ref{#1}}
\def\1{\bm{1}}

\DeclareMathAlphabet{\mathsfit}{\encodingdefault}{\sfdefault}{m}{sl}
\SetMathAlphabet{\mathsfit}{bold}{\encodingdefault}{\sfdefault}{bx}{n}

\usepackage{diagbox}
\usepackage[utf8]{inputenc} % allow utf-8 input
\usepackage[T1]{fontenc}    % use 8-bit T1 fonts
\usepackage{hyperref}       % hyperlinks
\usepackage{url}            % simple URL typesetting
\usepackage{booktabs}       % professional-quality tables
\usepackage{amsfonts}       % blackboard math symbols
\usepackage{nicefrac}       % compact symbols for 1/2, etc.
\usepackage{microtype}      % microtypography
\usepackage{xcolor}         % colors
\usepackage{amsmath}
\usepackage{ragged2e}
\usepackage{enumitem}
\usepackage{threeparttable}

\usepackage{float}
\usepackage{booktabs}
\usepackage{graphicx}

\usepackage{float} 
\usepackage{colortbl}
\usepackage{capt-of}
\usepackage{multirow}
\usepackage[most]{tcolorbox}
\usepackage{capt-of}
\tcbuselibrary{breakable, skins}
\usepackage{changepage}
\usepackage{caption}  % 提供 \captionof
\usepackage[export]{adjustbox}  % 导言区添加
\usepackage{lineno}
\usepackage{wrapfig}

\title{EG-VQA: Benchmarking Verifiable Video Question Answering with Grounded Temporal Evidence}

\author{
Linpeng Huang$^{1*}$ \quad
Weixing Chen$^{1*}$ \quad
Zexin Chen$^{1}$ \quad
Yang Liu$^{1,3\dagger}$ \quad
Liang Lin$^{1,2,3}$
\\[2pt]
$^{1}$Sun Yat-sen University
\quad
$^{2}$Peng Cheng Laboratory
\quad
$^{3}$X-Era AI Lab
\\[2pt]
$^{*}$Equal contribution
\qquad
$^{\dagger}$Corresponding Author
\\[2pt]
\texttt{\{huanglp33,chenwx228,chenzx276\}@mail2.sysu.edu.cn}
\\
\texttt{liuy856@mail.sysu.edu.cn, linliang@ieee.org}
\\[2pt]
\url{https://hcplab-sysu.github.io/EG-VQA/}
}

\begin{document}

\maketitle

\vspace{-5pt}
\begin{figure}[ht]
    \centering
    \vspace{-10pt}
    \includegraphics[width=1.0\linewidth]{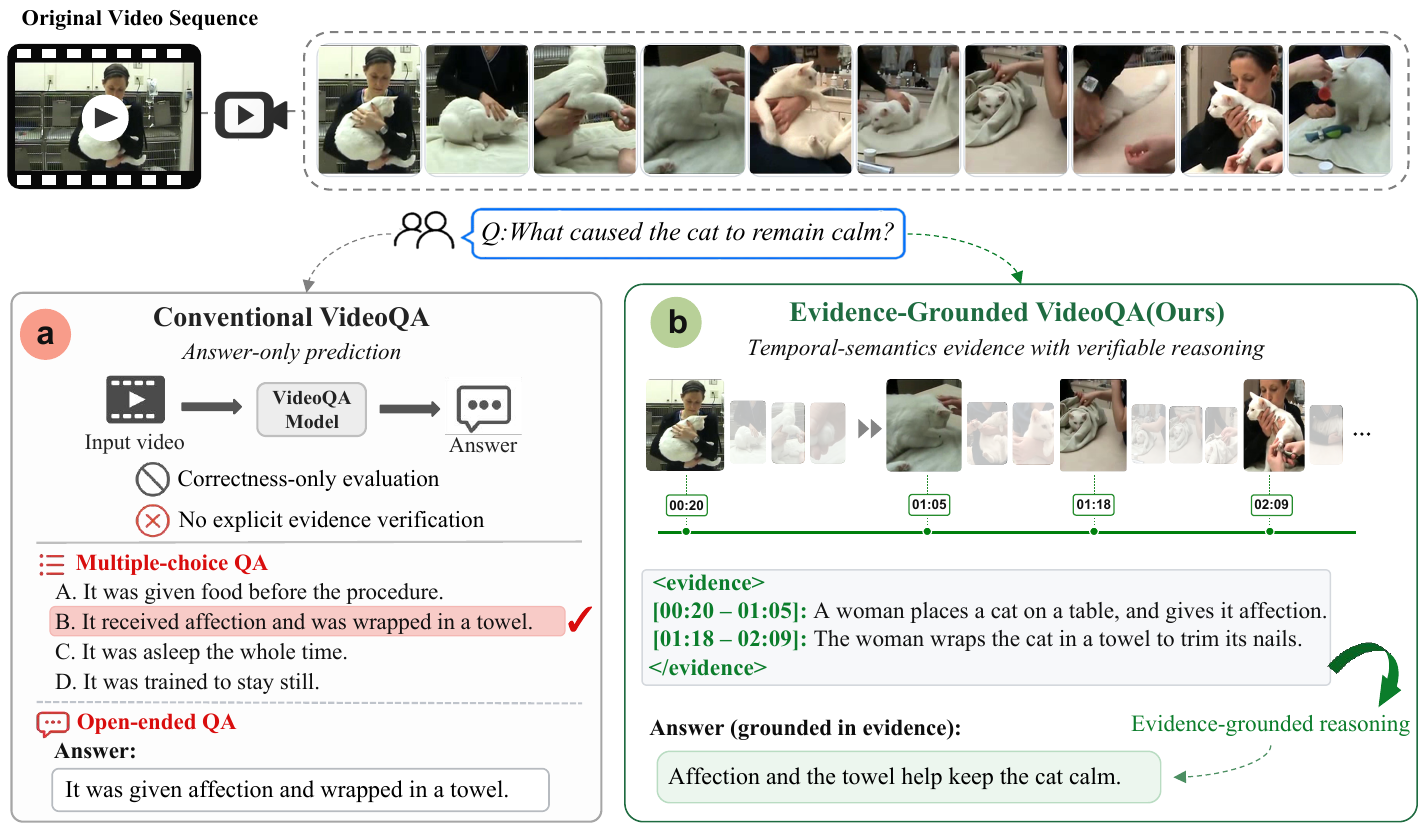}
    \vspace{-15pt}
    \caption{Overview of evidence-grounded VideoQA evaluation in EG-VQA. (a) Conventional VideoQA evaluation primarily focuses on answer prediction, without explicitly verifying whether the prediction is supported by relevant video evidence. (b) EG-VQA requires models to generate answers together with temporally localized textual evidence, enabling joint evaluation of answer correctness and evidence faithfulness.}
    \label{fig:eg_vqa_framework}
\end{figure}

\begin{abstract}
Recent advances in Video Large Language Models (Video-LLMs) have yielded promising performance on Video Question Answering (VideoQA). Nevertheless, existing benchmarks are predominantly evaluated through answer correctness, while the faithfulness of predicted evidence supporting those answers remains insufficiently evaluated. This disconnect between answer generation and evidence verification motivates the construction of the \textbf{Evidence-Grounded Video Question Answering Benchmark (EG-VQA)}, a large-scale open-ended benchmark in which each QA pair is annotated with temporally localized textual evidence, requiring models to jointly produce answers and verifiable evidence. EG-VQA comprises 2,067 videos and 11,838 QA pairs with fine-grained evidence annotations. To evaluate predicted evidence, \textbf{Evidence-Grounded F1 (EG-F1)} is introduced as a metric that jointly measures temporal alignment and semantic consistency between predicted and ground-truth evidence. Experiments reveal a substantial discrepancy between answer correctness and evidence faithfulness: even strong proprietary models can achieve high answer accuracy while exhibiting gaps in temporal-semantic grounding. To investigate evidence-aware learning under EG-VQA, we develop \textbf{EG-Reasoner}, an evidence-aware model trained with explicit evidence supervision. EG-Reasoner achieves strong evidence grounding performance among open-source models while maintaining competitive answer generation results compared with proprietary systems. These findings highlight the importance of explicitly modeling evidence grounding and suggest that evidence-aware supervision provides an effective direction toward more reliable and interpretable VideoQA systems.
% These findings suggest that stronger model scaling alone may not fully address evidence-grounded video understanding, and that evidence-aware supervision provides an effective direction toward more reliable and interpretable VideoQA systems.
\end{abstract}

%----------------------------------1 .Introduction-----------------------------
%------------------------------------------------------------------------------
\vspace{-8pt}
\section{Introduction}
\vspace{-3pt}

Video Question Answering (VideoQA) serves as a critical testbed for evaluating a model's capacity to comprehend dynamic visual content and reason over temporal events~\citep{zhong2022video,lin2026multimodal,jiang2025beyond}. Despite recent advances in large multimodal models, a fundamental limitation persists: current benchmarks often fail to distinguish between genuine video understanding and the exploitation of superficial correlations~\citep{ko2023large}.

Many widely used VideoQA datasets are constructed in a multiple-choice format, wherein models select answers from predefined options~\citep{tapaswi2016movieqa,Xiao2021NExTQANP,Wu2024STARAB,fu2025video}. Although this paradigm simplifies evaluation, it inadvertently introduces option bias, allowing models to rely on linguistic shortcuts rather than grounding predictions in visual evidence~\citep{liu2023cross,chen2025cross,Liu2023CausalVLRAT}, as shown in Figure~\ref{fig:eg_vqa_framework} (a). Furthermore, fine-grained evidence annotations are typically absent from these benchmarks, rendering it difficult to verify whether relevant video segments support a given prediction. 
Consequently, models frequently produce black-box answers devoid of interpretable reasoning~\citep{liu2022causal}. Even in open-ended settings, evaluation is conventionally restricted to answer correctness, while attention to correct temporal regions is overlooked~\citep{xiao2024can,lei2020tvqa+}. These limitations underscore a critical gap: correct answers do not necessarily imply that the underlying reasoning is faithful to the video content. For VideoQA systems to be reliable, both the answer and its provenance within the video should be explicitly verifiable~\citep{xiao2024can}.

Motivated by this challenge, we introduce the \textbf{Evidence-Grounded Video Question Answering Benchmark (EG-VQA)}, a large-scale open-ended VideoQA benchmark designed to jointly evaluate answer correctness and evidence faithfulness, i.e., whether the predicted evidence temporally and semantically supports the answer. Unlike conventional answer-only evaluation, EG-VQA requires models to generate both an answer and its supporting temporally localized evidence, enabling direct verification of whether predictions are grounded in the observed video content. As illustrated in Figure~\ref{fig:eg_vqa_framework}(b), each evidence unit is represented as a temporal interval paired with a textual event description, allowing evaluation of not only where relevant information appears but also whether the described evidence semantically supports the answer. The resulting dataset contains 2,067 videos and 11,838 QA pairs spanning four reasoning categories: descriptive, temporal, causal, and counterfactual.

Extensive evaluation of state-of-the-art multimodal models on EG-VQA reveals a substantial gap between answer correctness and evidence faithfulness. While strong models can often generate correct answers, their predicted evidence may still fail to accurately localize and describe the supporting video events. This discrepancy indicates that answer accuracy alone is insufficient for evaluating reliable VideoQA systems and motivates the study of evidence-aware learning strategies under the proposed benchmark.

To further study evidence-grounded VideoQA under EG-VQA, we develop \textbf{EG-Reasoner} as an evidence-aware reference model trained with explicit evidence supervision. EG-Reasoner generates answers together with temporally localized evidence, providing a model-level investigation of how evidence-aware supervision affects VideoQA performance. In addition, we introduce \textbf{Evidence-Grounded F1 (EG-F1)}, a metric tailored to the temporal-semantic evidence formulation of EG-VQA, which jointly evaluates temporal alignment and semantic consistency between predicted and reference evidence through optimal bipartite matching. Experiments show that evidence-aware optimization substantially improves evidence grounding performance and achieves competitive results among open-source Video-LLMs.
Our contributions are summarized as follows:

\begin{itemize}[itemsep=2pt, topsep=2pt, parsep=1pt, partopsep=0pt]
\item We introduce \textbf{EG-VQA}, a large-scale open-ended VideoQA benchmark that associates each QA pair with temporally localized textual evidence. By representing evidence as temporal intervals paired with event descriptions, EG-VQA enables joint evaluation of answer correctness and evidence faithfulness.

\item We introduce \textbf{Evidence-Grounded F1 (EG-F1)}, a metric designed for temporal-semantic evidence evaluation, which jointly measures temporal alignment and semantic consistency between predicted and reference evidence through optimal bipartite matching.

\item We develop \textbf{EG-Reasoner} as an evidence-aware reference model to investigate grounding-oriented learning under EG-VQA. Through explicit evidence supervision, EG-Reasoner generates answers jointly with temporally localized evidence and demonstrates the effectiveness of evidence-aware optimization.
     
\end{itemize}

%-------------------------2 .Related Work-----------------------------------
%---------------------------------------------------------------------------
\vspace{-5pt}
\section{Related Work}
\vspace{-3pt}
\label{sec:related_work}

\subsection{Video Question Answering Benchmarks}
\vspace{-5pt}

Video Question Answering (VideoQA) benchmarks have evolved from short-video understanding toward increasingly complex temporal and causal reasoning. Early datasets such as TGIF-QA~\citep{jang2017tgif}, MovieQA~\citep{tapaswi2016movieqa}, and MSRVTT-QA~\citep{xu2017video} established VideoQA settings over GIFs, movies, and short web videos, while later benchmarks including ActivityNet-QA~\citep{yu2019activitynet}, TVQA~\citep{lei2018tvqa}, NExT-QA~\citep{Xiao2021NExTQANP}, STAR~\citep{Wu2024STARAB}, and Video-MME~\citep{fu2025video} expanded evaluation to longer videos and more diverse temporal and compositional reasoning. Many widely used benchmarks adopt multiple-choice protocols, which facilitate automatic evaluation but may introduce option-related shortcuts and do not by themselves establish whether an answer is supported by the relevant video content~\citep{chen2025cross,liu2023cross,wei2023visual,liu2022causal}.

To explicitly verify visual support for predicted answers, grounded VideoQA benchmarks augment answer supervision with evidence localization. TVQA+~\citep{lei2020tvqa+} introduces spatio-temporal grounding, while NExT-GQA~\citep{xiao2024can} associates QA pairs with temporal labels for answer-supporting moments. MultiHop-EgoQA~\citep{Chen2024GroundedMV} further requires models to answer questions while localizing multiple relevant temporal intervals in long-form egocentric videos, and TOGA~\citep{Gupta2025TOGATG} studies temporally grounded open-ended VideoQA. HERBench~\citep{ben2026herbench} focuses on the complementary challenge of integrating multiple temporally dispersed visual cues. Building on this direction, EG-VQA represents each supporting evidence unit as a temporal-semantic pair, consisting of a temporal span and a standalone textual description of the corresponding event. This requires models to predict not only where the supporting evidence occurs but also whether the evidence description semantically supports the answer, enabling direct verification of both temporal alignment and semantic fidelity.

\vspace{-5pt}
\subsection{Video-LLMs and Evidence-Grounded Reasoning}
\vspace{-5pt}

Recent Video Large Language Models (Video-LLMs) and multimodal large language models have substantially advanced video understanding through improved visual perception, temporal modeling, and language generation. Representative model families such as Qwen-VL~\citep{Qwen2.5-VL,bai2025qwen3}, InternVL~\citep{zhu2025internvl3,wang2025internvl3}, and VideoChat~\citep{Li2023VideoChatCV} demonstrate strong capabilities across video question answering, dialogue, and long-video understanding. More recent reasoning-oriented methods further employ post-training to improve temporal and spatio-temporal reasoning. Time-R1~\citep{wang2026time} applies reinforcement-learning-based post-training for temporal video grounding, while VideoChat-R1~\citep{li2025videochat} uses reinforcement fine-tuning with temporally aware rewards to improve spatio-temporal perception and reasoning.

These methods primarily optimize answer generation, temporal reasoning, or general video understanding objectives. EG-Reasoner instead focuses on generating explicit temporal-semantic supporting evidence together with the final answer, with evidence supervision defined over the same interval-description representation used by EG-VQA. This allows training to directly account for both the temporal and semantic quality of predicted supporting evidence.

%-----------3 .Evidence-grounded video question answering benchmark---------
%---------------------------------------------------------------------------
\vspace{-5pt}
\section{Evidence-Grounded Video Question Answering Benchmark}
\begin{figure}[t]
    \centering
    \includegraphics[
        width=1.0\linewidth,
        %trim=左 下 右 上
        %trim=0pt 210pt 60pt 0pt,
        clip
    ]{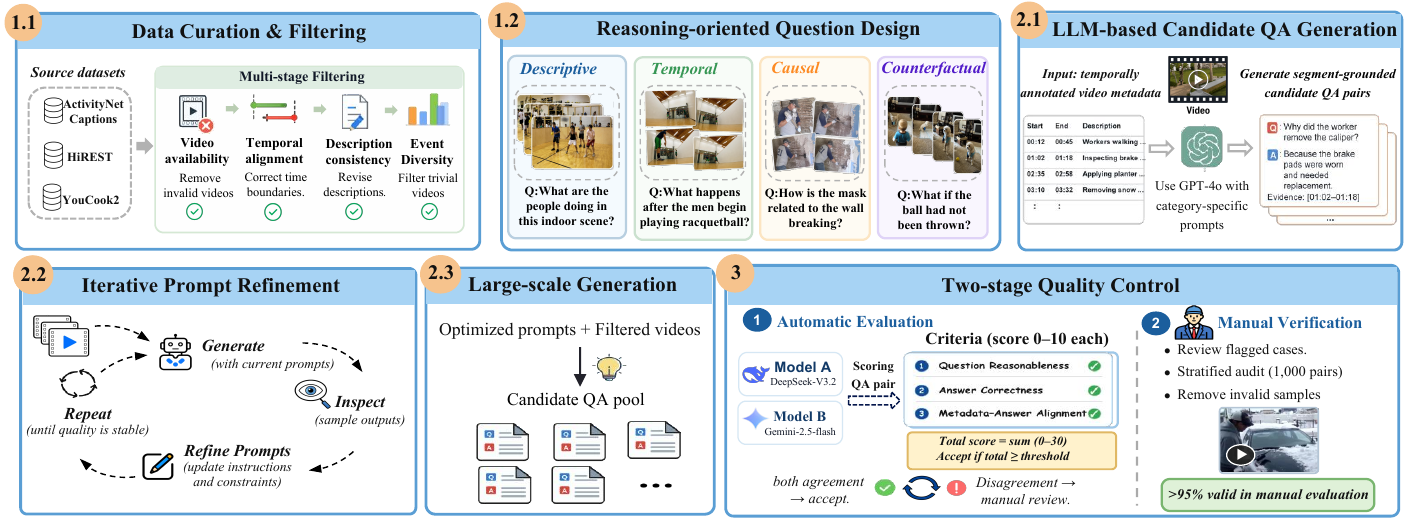}
      \vspace{-15pt}
    \caption{Overview of the EG-VQA dataset construction pipeline. The pipeline includes data curation and filtering, reasoning-oriented QA generation with prompt refinement, and two-stage quality control.}
    \label{fig:dataset_pipeline}
    % \vspace{-10pt}
\end{figure}

\vspace{-3pt}
This section describes the construction pipeline of the Evidence-Grounded Video Question Answering (EG-VQA) Benchmark, including the data curation, reasoning-oriented QA generation with prompt refinement, and two-stage quality control, as shown in Figure~\ref{fig:dataset_pipeline}. More details discussed in \ref{app:construction}.
\vspace{-0.5em}
\subsection{Data Curation and Question Design}
\vspace{-0.5em}
\label{sec:curation}

The foundation of EG-VQA is built from three existing video sources: ActivityNet Captions~\citep{caba2015activitynet}, HiREST~\citep{zala2023hierarchical}, and YouCook2~\citep{zhou2018towards}, which provide temporally segmented event annotations across diverse video domains, including open-domain activities, instructional procedures, and long-form event-centric videos. Specifically, we use the official validation splits of ActivityNet Captions and YouCook2, and the validation and test splits of HiREST as the source pool. These source partitions are used to obtain videos and temporal annotations, while EG-VQA subsequently constructs its own train–test partition at the video level.
The raw annotations are standardized into a unified format in which each video is represented as an ordered sequence of timestamped segments paired with concise textual event descriptions. Because the original segment boundaries and descriptions occasionally deviate from the actual video content, we remove corrupted or inaccessible clips, correct misaligned temporal boundaries, revise inconsistent descriptions, and exclude videos with low event diversity that would otherwise yield trivial questions. After curation, 2,067 videos remain, with the corrected temporal descriptions serving as the evidence backbone. Detailed filtering criteria and source-wise statistics are provided in Appendix~\ref{app:filtering} and ~\ref{app:statistics}.

EG-VQA covers four reasoning categories. Descriptive (Desc.) questions focus on directly observable actions, objects, and states; Temporal (Temp.) questions require understanding event order and procedural sequences; Causal (Caus.) questions involve reasoning about event dependencies and consequences; and Counterfactual (Cntrf.) questions require reasoning over hypothetical alterations of observed events. Together, these categories span direct perception, temporal reasoning, causal inference, and counterfactual reasoning grounded in specific video events.

\vspace{-0.5em}
\subsection{Dataset Construction and Quality Control}
\vspace{-0.5em}
\label{sec:construction}
QA pairs are generated through a human-in-the-loop pipeline that iterates between automated generation and manual refinement. Given the temporally annotated video metadata, GPT-4o is prompted to generate candidate questions, answers, and supporting evidence for each reasoning category. The initial prompts are applied to a held-out subset and manually inspected for common failure modes, including video-independent questions, weak grounding to the metadata, and answers that omit necessary temporal context. The prompts are iteratively refined until the generated samples consistently satisfy the intended quality requirements, after which they are deployed across the curated video pool. Prompt templates and refinement details are provided in Appendix~\ref{app:prompt_design}.

We then apply a two-stage quality-control protocol to all candidate QA pairs. First, DeepSeek-V3.2 and Gemini-2.5-Flash independently evaluate each sample for question reasonableness, answer correctness, and evidence alignment using a structured scoring rubric. Samples receiving consistent accept/reject decisions are processed accordingly, while model disagreements are sent to human annotators for adjudication. After this full-dataset screening, we conduct an additional stratified random audit of 1,000 QA pairs, balanced across data sources, video durations, and question types. More than 95\% of the audited samples are verified as valid. Minor, correctable issues are manually revised and retained, whereas substantively invalid samples are removed. The complete validation protocol is provided in Appendix~\ref{app:quality_control}.
% Follow-up inspection of the observed failure patterns does not reveal additional systematic error modes. The complete validation protocol is provided in Appendix~\ref{app:quality_control}.

Finally, we construct the train–test partition using a stratified video-level split, controlling for source dataset, video duration, and evidence density to maintain comparable distributions across the two partitions. All QA pairs and evidence annotations derived from the same video are assigned to the same partition, ensuring that no video appears in both the training and test sets. The resulting dataset contains 1,731 training videos with 8,949 QA pairs and 336 test videos with 2,889 QA pairs. Each question is associated with an average of 2.27 supporting evidence segments. Detailed source-wise and per-split statistics are reported in Appendix~\ref{app:statistics}.

\vspace{-0.5em}
\subsection{Evaluation via Evidence-Grounded F1}
\vspace{-0.5em}
\label{sec:egf1}
In EG-VQA, evidence quality depends on both accurate temporal localization and semantic consistency with the supporting video events. Existing evaluation protocols often assess temporal overlap and textual similarity separately, allowing a prediction to perform well on one dimension while failing on the other. EG-F1 therefore evaluates predicted evidence as temporal-semantic units, requiring matched evidence to align with the reference in both time and meaning.

Formally, let the ground-truth evidence set be $\mathcal{G} = \{g_i\}_{i=1}^{|\mathcal{G}|}$ and the predicted set be $\mathcal{P} = \{p_j\}_{j=1}^{|\mathcal{P}|}$, where each item $g_i = (t_i, d_i)$ and $p_j = (t_j, d_j)$ consists of a temporal segment and a textual description. The metric proceeds by computing pairwise temporal overlap $\mathrm{IoU}_{ij}$ and semantic similarity $\mathrm{Sim}_{ij}$ between all pairs, retaining only those edges that satisfy thresholds on both dimensions, and then finding an optimal one-to-one matching via the Hungarian algorithm~\citep{kuhn1955hungarian}. Precision and recall are derived from the number of matched pairs $M$, yielding
\begin{equation}
    \mathrm{EG\!-\!F1} = \frac{2 \cdot (M/|\mathcal{P}|) \cdot (M/|\mathcal{G}|)}{(M/|\mathcal{P}|) + (M/|\mathcal{G}|)}.
\end{equation}
The complete algorithmic formulation, including threshold settings and the soft variant used during training, is deferred to Appendix~\ref{app:egf1_details}.

\label{sec:benchmark}

%-----------------------------------4 .EG-Reasoner--------------------------
%---------------------------------------------------------------------------
% \vspace{-5pt}
\section{EG-Reasoner: An Evidence-Aware Baseline}
\label{sec:method}

To investigate how the evidence supervision provided by EG-VQA can be incorporated into model post-training, we develop EG-Reasoner as an evidence-aware baseline. EG-Reasoner generates structured responses containing supporting evidence, an intermediate reasoning trace, and a final answer. We adopt standard Group Relative Policy Optimization (GRPO)~\cite{shao2024deepseekmath,guo2025deepseek} for post-training, while the task-specific design lies in a composite reward that evaluates output format, answer correctness, and temporal-semantic evidence quality. Figure~\ref{fig:eg_reasoner_framework} provides an overview of the training framework.

\subsection{Structured Generation Formulation}
\label{sec:formulation}

\begin{figure}[t]
                \centering
    \includegraphics[width=1.0\linewidth]{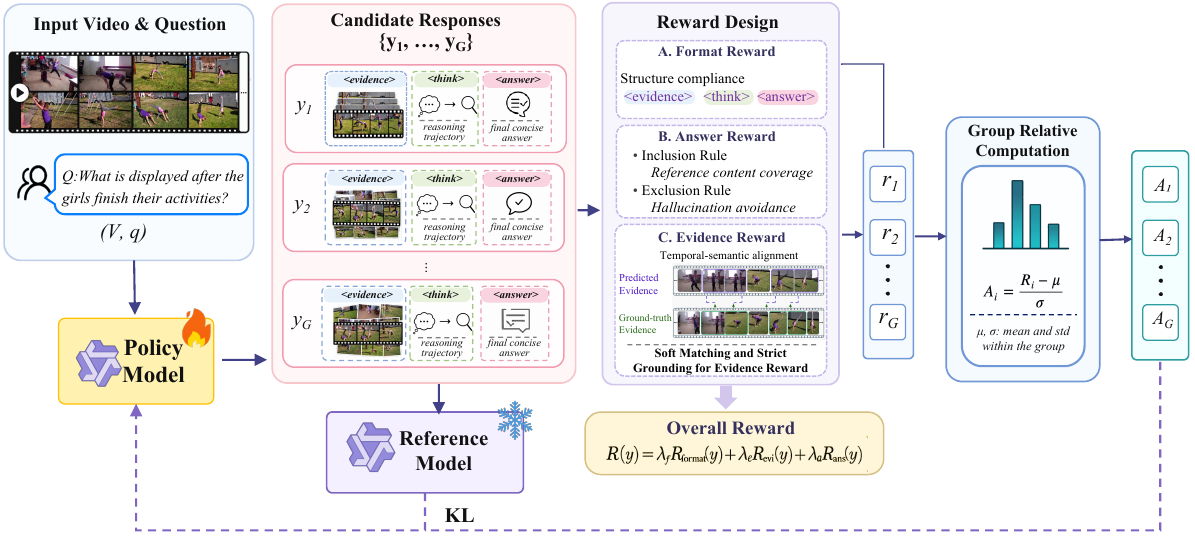}
                  \vspace{-15pt}
                \caption{Overview of the EG-Reasoner training framework. Given a video-question pair, the model generates structured responses containing evidence, intermediate reasoning, and answer, and is optimized with GRPO using a composite reward for format compliance, answer correctness, and evidence faithfulness.}
                \label{fig:eg_reasoner_framework}
                % \vspace{-10pt}
            \end{figure}

Given an input pair $(V, q)$, the model generates a structured response
\begin{equation}
    y = (\hat{\mathcal{E}}, \hat{\mathcal{T}}, \hat{\mathcal{A}}),
\end{equation}
where $\hat{\mathcal{E}} = \{(\hat{t}_k, \hat{d}_k)\}$ is a set of predicted evidence segments with temporal boundaries and textual descriptions, $\hat{\mathcal{T}}$ is a reasoning trajectory that chains the evidence into an explanatory narrative, and $\hat{\mathcal{A}}$ is a concise final answer. This three-block design, shown in the candidate responses of Figure~\ref{fig:eg_reasoner_framework}, separates the predicted supporting evidence from the final answer, enabling the grounding quality of the response to be evaluated directly rather than inferred solely from answer correctness.

\subsection{Composite Reward Design}
\label{sec:reward}

Each sampled response is evaluated using a composite reward:
\begin{equation}
    R(y) = \lambda_f R_{\mathrm{format}}(y) + \lambda_e R_{\mathrm{evi}}(y) + \lambda_a R_{\mathrm{ans}}(y).
\end{equation}
where \(R_{\mathrm{format}}\), \(R_{\mathrm{ans}}\), and \(R_{\mathrm{evi}}\) measure structure compliance, answer correctness, and evidence quality, respectively. The corresponding weights are provided in Appendix~\ref{app:hyperparams}.

\textbf{Format reward.} The format reward encourages responses to follow the predefined output schema containing the \textless evidence\textgreater, \textless think\textgreater, and \textless answer\textgreater\ blocks. It assigns a binary score according to whether the required structure is correctly produced.

\textbf{Answer reward.} The answer reward evaluates whether the generated answer covers the key content of the reference answer while avoiding hallucinated or unsupported statements. A rule-guided LLM evaluator follows two complementary principles. The inclusion principle rewards coverage of relevant reference content, while the exclusion principle penalizes information that contradicts or cannot be supported by the video metadata. Partially correct answers receive intermediate credit to provide a denser optimization signal. Detailed scoring rules are provided in Appendix~\ref{app:answer_reward}.

\textbf{Evidence reward.} Directly using the thresholded EG-F1 as a training reward produces sparse supervision, since early predictions rarely satisfy both the temporal and semantic thresholds required for a valid evidence match. We therefore introduce a soft matching variant that assigns partial credit according to the product of temporal overlap and semantic similarity between predicted and reference evidence. Optimal bipartite matching is performed without hard thresholding to obtain a dense evidence score.

Soft matching alone, however, may reward loosely aligned evidence. We therefore combine the soft score with the original thresholded EG-F1:
\begin{equation}
    R_{\mathrm{evi}} = R_{\mathrm{soft}} + R_{\mathrm{EG\!-\!F1}}.
\end{equation}
The soft component provides graded feedback for partially aligned evidence, while the hard component preserves the stricter temporal-semantic criterion used by EG-F1. The complete formulation of the soft variant is provided in Appendix~\ref{app:soft_variant}.

% \vspace{-0.5em}
\subsection{Optimization with GRPO}
\vspace{-0.3em}
\label{sec:grpo}
A structured response may admit multiple semantically valid realizations. Correct answers can be expressed through different phrasings, while supporting events may admit different but valid textual descriptions. A single-reference supervised objective observes only one annotated realization and optimizes token-level likelihood against that sequence, rather than directly optimizing task-level answer correctness and set-level evidence quality.

Given the composite reward defined above, we optimize EG-Reasoner using standard GRPO. For each video-question pair, the policy samples a group of candidate responses, each of which is evaluated by the composite reward. Their relative rewards are normalized within the group to estimate advantages, and the policy is updated to increase the likelihood of higher-quality responses. A KL regularization term constrains the policy from deviating excessively from the reference policy. The full GRPO objective, including group-relative advantage computation, clipping, and KL regularization, is provided in Appendix~\ref{app:grpo}. Additional implementation details and hyperparameters are reported in Appendix~\ref{app:hyperparams}.

% During training, the visual backbone remains frozen and only the language-model parameters are optimized. Additional implementation details and hyperparameters are reported in Appendix~\ref{app:hyperparams}.

%-----------------------------------5.Experiments---------------------------
%---------------------------------------------------------------------------
\vspace{-5pt}
\section{Experiments}
\label{others_5}

%---------------------------------------------------------------------------
\vspace{-5pt}
\subsection{Experimental Setup}
\vspace{-5pt}

\textbf{Models.}
We evaluate a diverse set of Video-LLMs spanning proprietary multimodal models, general-purpose open-source models, and reasoning-oriented video models. The evaluated systems cover representative model families with different scales and training paradigms. Detailed descriptions of all baselines are provided in Appendix~\ref{app:baseline_details}.

\textbf{Evaluation Metrics.}
We evaluate model performance in terms of answer correctness and evidence grounding. For open-ended answer evaluation, we use Gemini-2.5-Pro as an LLM judge and report strict and relaxed accuracy. On a 200-sample human audit, the judge achieves 96.5\% exact agreement with human evaluation (Cohen's $\kappa=0.944$). For evidence grounding, we report event-level F1 under multiple temporal IoU thresholds together with EG-F1. We additionally report the Grounded Answer Score (GAS), which measures whether answer correctness and evidence grounding are achieved on the same samples. Detailed evaluation protocols are provided in Appendix~\ref{app:answer_eval} to ~\ref{app:gas}.

\textbf{Implementation details.}
We adopt Qwen2.5-VL-7B-Instruct as the base model of EG-Reasoner and fine-tune it on the EG-VQA training split.
During both training and evaluation, the model is required to generate structured outputs following the unified format defined in Section~4. 
Detailed training hyperparameters and evaluation configurations are provided in Appendix~\ref{app:hyperparams} and Appendix~\ref{app:experimental_setup}.

\textbf{Human study.} To estimate human performance on EG-VQA, we randomly sample 200 test QA pairs covering all four reasoning categories and ask three human annotators to provide open-ended answers together with supporting evidence under the same task setting as the evaluated models. Their responses are evaluated using the same answer and evidence metrics, and we report the average performance across annotators.
%-------------------------------------------------   --------------------------
\vspace{-5pt}
\subsection{Main Results}
\vspace{-5pt}

Table~\ref{tab:merged-experimental-evidence-results} summarizes the main results on EG-VQA. GPT-5.6 Luna achieves the highest overall answer accuracy among the evaluated models, with 32.12\% strict and 45.95\% relaxed accuracy, while remaining below human performance (56.67\% and 70.95\%). Among open-source models, EG-Reasoner performs best overall, reaching 26.88\% strict and 42.71\% relaxed accuracy, improving over Qwen2.5-VL-7B by 12.57 and 14.75 percentage points, respectively. Its performance is particularly strong on counterfactual questions, where it achieves 48.16\% strict and 60.50\% relaxed accuracy.  

\begin{table}[t]
\begin{center}
\begin{minipage}[t]{\linewidth}
\centering
% \vspace{-12pt}
\captionof{table}{Overall performance comparison on EG-VQA.}
\vspace{-10pt}
\label{tab:merged-experimental-evidence-results}
{\setlength{\tabcolsep}{2.6pt}
\renewcommand{\arraystretch}{1.1}
\resizebox{\linewidth}{!}{%
\begin{tabular}{l*{5}{c}*{5}{c}*{2}{c}c}
\toprule
\multirow{2}{*}{\textbf{Model}}
& \multicolumn{5}{c}{\textbf{Strict Accuracy}}
& \multicolumn{5}{c}{\textbf{Relaxed Accuracy}}
& \multicolumn{2}{c}{\textbf{Evidence Grounding}}
& \multicolumn{1}{c}{\textbf{Joint}} \\
\cmidrule(lr){2-6}
\cmidrule(lr){7-11}
\cmidrule(lr){12-13}
\cmidrule(lr){14-14}
& \textbf{Desc.}
& \textbf{Temp.}
& \textbf{Caus.}
& \textbf{Cntrf.}
& \textbf{Overall}
& \textbf{Desc.}
& \textbf{Temp.}
& \textbf{Caus.}
& \textbf{Cntrf.}
& \textbf{Overall}
& \shortstack{\textbf{Avg.~F1}}
& \shortstack{\textbf{Avg.~EG-F1}}
& \textbf{GAS} \\
\midrule

\textbf{Human Results*}
& 52.08 & 57.69 & 51.72 & 65.38 & 56.67
& 68.75 & 73.08 & 65.52 & 76.92 & 70.95
& 73.76 & 56.34 & 33.81 \\

\midrule
\rowcolor{gray!15}
\multicolumn{14}{c}{\textit{Commercial Large Multimodal Models}} \\
\midrule

GPT-4o~\cite{hurst2024gpt}
& 18.90 & 23.08 & 37.20 & 35.10 & 28.83
& 40.10 & 38.94 & 48.00 & 53.50 & 45.24
& 15.58 & 3.33 & 2.29 \\

Gemini-2.5-Flash~\cite{comanici2025gemini}
& 14.58 & 20.45 & 33.93 & 28.89 & 24.75
& 37.50 & 36.36 & 43.75 & 46.67 & 41.15
& 17.03 & 4.99 & 2.74 \\

GPT-5.6 Luna
& 27.08 & 21.00 & 39.58 & 40.40 & 32.12
& 43.23 & 35.00 & 47.92 & 57.58 & 45.95
& 32.79 & 10.56 & 6.81 \\

Qwen3.7 Plus
& 24.00 & 18.45 & 30.61 & 23.47 & 24.13
& 42.00 & 33.50 & 45.41 & 41.84 & 40.63
& 20.73 & 5.89 & 2.58 \\

\midrule
\rowcolor{gray!15}
\multicolumn{14}{c}{\textit{Open-sourced Video-LLMs}} \\
\midrule

Qwen2.5-VL-7B~\cite{Qwen2.5-VL}
& 9.91 & 7.73 & 22.36 & 16.80 & 14.31
& 24.96 & 20.07 & 36.21 & 30.30 & 27.96
& 13.86 & 3.33 & 1.80 \\

InternVL-3-8B~\cite{zhu2025internvl3}
& 9.71 & 5.24 & 18.11 & 13.64 & 11.71
& 27.91 & 18.75 & 30.45 & 28.72 & 26.36
& 9.48 & 2.02 & 1.01 \\

Time-R1-7B~\cite{wang2026time}
& 10.52 & 7.74 & 22.73 & 21.15 & 15.66
& 27.52 & 20.49 & 39.24 & 34.24 & 30.44
& 24.89 & 6.87 & 3.23 \\

VideoChat-R1-7B~\cite{li2025videochat}
& 13.24 & 9.99 & 24.10 & 19.81 & 16.87
& 30.29 & 22.90 & 41.37 & 34.15 & 32.21
& 27.00 & 7.65 & 3.95 \\

VideoChat-R1-Thinking-7B~\cite{li2025videochat}
& 9.91 & 8.40 & 24.50 & 25.65 & 17.29
& 26.75 & 20.97 & 40.09 & 38.74 & 31.76
& 21.56 & 5.76 & 2.86 \\

VideoChat-R1.5-7B~\cite{yan2026videochat}
& 10.65 & 8.12 & 22.92 & 19.02 & 15.29
& 28.46 & 22.04 & 39.01 & 32.01 & 30.42
& 17.29 & 5.08 & 2.41 \\

TW-GRPO~\cite{dang2025reinforcing}
& 12.37 & 7.48 & 25.80 & 24.21 & 17.58
& 27.79 & 19.56 & 39.97 & 36.70 & 31.14
& 16.17 & 3.77 & 2.04 \\

\midrule
\textbf{EG-Reasoner (Ours)}
& \textbf{14.20} & \textbf{10.92} & \textbf{33.16} & \textbf{48.16} & \textbf{26.88}
& \textbf{34.12} & \textbf{27.10} & \textbf{48.46} & \textbf{60.50} & \textbf{42.71}
& \textbf{37.89} & \textbf{12.28} & \textbf{7.14} \\

\bottomrule
\end{tabular}%
}
}

\vspace{-1mm}

{\footnotesize
\justifying
\textit{Note.}
Avg. F1 and Avg. EG-F1 denote averages across the threshold settings specified in Appendix~\ref{app:evidence_eval}.\par
}

\end{minipage}
\end{center}
% \vspace{-15pt}

\end{table}

EG-Reasoner shows a stronger advantage in evidence grounding, achieving 37.89 Avg. F1 and 12.28 Avg. EG-F1, outperforming GPT-5.6 Luna (32.79 and 10.56) and VideoChat-R1-7B (27.00 and 7.65). It also obtains the highest model GAS of 7.14, indicating stronger sample-level co-occurrence of correct answers and grounded evidence. Human performance remains higher overall, indicating substantial room for improvement in model performance. Detailed threshold-specific grounding results are provided in Appendix~\ref{app:full_evidence_results}, and additional analyses of video dependency and the answer correctness–evidence grounding relationship are provided in Appendix~\ref{app:benchmark_diagnosis}.
\vspace{-5pt}
\subsection{Ablation Studies}
\vspace{-5pt}
\label{sec:ablation}
We perform controlled ablations to analyze the contribution of different training components in EG-Reasoner. Results are summarized in Table~\ref{tab:evidence_training_ablation}.
\vspace{-5pt}
\begin{table}[ht]
\centering
\caption{Ablation study of evidence-aware training on EG-VQA.}
\vspace{-10pt}
\label{tab:evidence_training_ablation}

\small
\setlength{\tabcolsep}{4.5pt}
\renewcommand{\arraystretch}{1.0}

\begin{tabular}{lcccc}
\toprule
\textbf{Model}
& \textbf{Strict Acc.$\uparrow$}
& \textbf{Relaxed Acc.$\uparrow$}
& \textbf{Avg. F1$\uparrow$}
& \textbf{Avg. EG-F1$\uparrow$}
\\
\midrule

Qwen2.5-VL-7B
& 14.31
& 27.96
& 13.86
& 3.33
\\
\rowcolor{gray!15}
\quad + SFT
& 13.12
& 24.47
& 20.29
& 4.85
\\
\rowcolor{gray!15}
\quad + RL w/o Evidence Reward
& 22.21
& 36.17
& 15.81
& 4.21
\\
\rowcolor{gray!15}
\quad + RL w/o Soft Evidence Reward
& 24.00
& 37.96
& 15.33
& 3.62
\\
\rowcolor{gray!15}
\textbf{\quad + RL w/ Evidence Reward (EG-Reasoner)}
& \textbf{26.88}
& \textbf{42.71}
& \textbf{37.89}
& \textbf{12.28}
\\

\midrule
VideoChat-R1-7B
& 16.87
& 32.21
& 27.00
& 7.65
\\
\rowcolor{gray!15}
\quad + RL w/o Evidence Reward
& 25.42
& 40.26
& 22.07
& 6.51
\\

\bottomrule
\end{tabular}
\end{table}
\vspace{-5pt}

\textbf{Effect of supervised fine-tuning.} SFT improves evidence grounding quality, increasing Avg. F1 from 13.86 to 20.29 and Avg. EG-F1 from 3.33 to 4.85, while slightly reducing answer accuracy. This suggests that SFT improves structured evidence prediction, but token-level optimization on reference responses does not fully align with the temporal-semantic grounding objective.

\textbf{Effect of reinforcement learning without evidence rewards.} RL without evidence rewards improves answer generation, increasing relaxed accuracy from 27.96\% to 36.17\%. However, evidence grounding improves only marginally (Avg. EG-F1: 3.33 → 4.21), indicating that optimizing response-level objectives alone is insufficient for precise evidence grounding.

\textbf{Effect of evidence-aware rewards.} Introducing evidence-aware rewards leads to substantial improvements in evidence grounding compared with RL without evidence rewards, demonstrating that explicitly optimizing evidence quality better aligns training with the evaluation objective. Removing the soft evidence component further degrades grounding performance, highlighting the importance of continuous temporal-semantic supervision. Moreover, VideoChat-R1-7B trained with the same reinforcement learning procedure but without evidence rewards still underperforms, suggesting that access to EG-VQA training data alone is insufficient and that evidence-aware supervision is essential for grounded video reasoning.
%---------------------------------------------------------------------------
\vspace{-5pt}
\subsection{Zero-Shot Generalization Beyond EG-VQA}
\vspace{-5pt}
\label{sec:generalization}
To evaluate the transferability of evidence-aware training beyond the original benchmark, we conduct zero-shot evaluation on two existing video reasoning benchmarks, NExT-GQA and ReXTime. No additional training or task-specific adaptation is performed on these benchmarks. For ReXTime, we exclude all videos that appear in the EG-VQA training split, along with their associated QA pairs, before evaluation. Results are summarized in Table~\ref{tab:cross_benchmark}.

\vspace{-5pt}
\begin{table}[ht]
\centering
\caption{Zero-shot transfer results on NExT-GQA and ReXTime.}
\vspace{-10pt}
\label{tab:cross_benchmark}

\small
\setlength{\tabcolsep}{7.5pt}
\renewcommand{\arraystretch}{1.0}

\begin{tabular}{lcccccc}
\toprule
& \multicolumn{3}{c}{\textbf{NExT-GQA}~\citep{xiao2024can}}
& \multicolumn{3}{c}{\textbf{ReXTime}~\citep{chen2024rextime}}
\\
\cmidrule(lr){2-4}
\cmidrule(lr){5-7}
\textbf{Method}
& \textbf{Acc}
& \textbf{mIoU}
& \textbf{Acc@GQA}
& \textbf{Acc}
& \textbf{mIoU}
& \textbf{Acc@GQA}
\\
\midrule

Qwen2.5-VL-7B~\cite{Qwen2.5-VL}
& 72.7
& 16.6
& 11.1
& 60.4
& 14.6
& 9.6
\\

Time-R1-7B~\cite{wang2026time}
& 78.4
& 28.6
& 19.4
& 64.3
& 25.4
& 17.9
\\

VideoChat-R1-7B~\cite{li2025videochat}
& 75.6
& 30.8
& 20.7
& 63.6
& 29.1
& 21.4
\\

TW-GRPO~\cite{dang2025reinforcing}
& 74.0
& 17.4
& 10.5
& 60.8
& 16.0
& 11.4
\\

VideoChat-R1.5~\cite{yan2026videochat}
& 77.9
& 15.4
& 9.1
& \textbf{65.7}
& 12.2
& 6.3
\\

\textbf{EG-Reasoner (Ours)}
& \textbf{78.7}
& \textbf{32.6}
& \textbf{23.3}
& 64.8
& \textbf{34.4}
& \textbf{25.5}
\\

\bottomrule
\end{tabular}
\end{table}
\vspace{-5pt}

EG-Reasoner demonstrates strong transferability across both benchmarks. On NExT-GQA, it achieves the best overall performance, leading both answer accuracy and temporal grounding metrics among all evaluated models. On ReXTime, while VideoChat-R1.5 obtains slightly higher answer accuracy, EG-Reasoner achieves the strongest temporal localization performance, obtaining the highest mIoU and Acc@GQA. These results indicate that evidence-aware training improves general temporal reasoning and grounding capabilities beyond the original EG-VQA setting.

\vspace{-5pt}
\subsection{Case Study}
\begin{figure}[ht]
    \centering
    \includegraphics[width=1.0\linewidth]{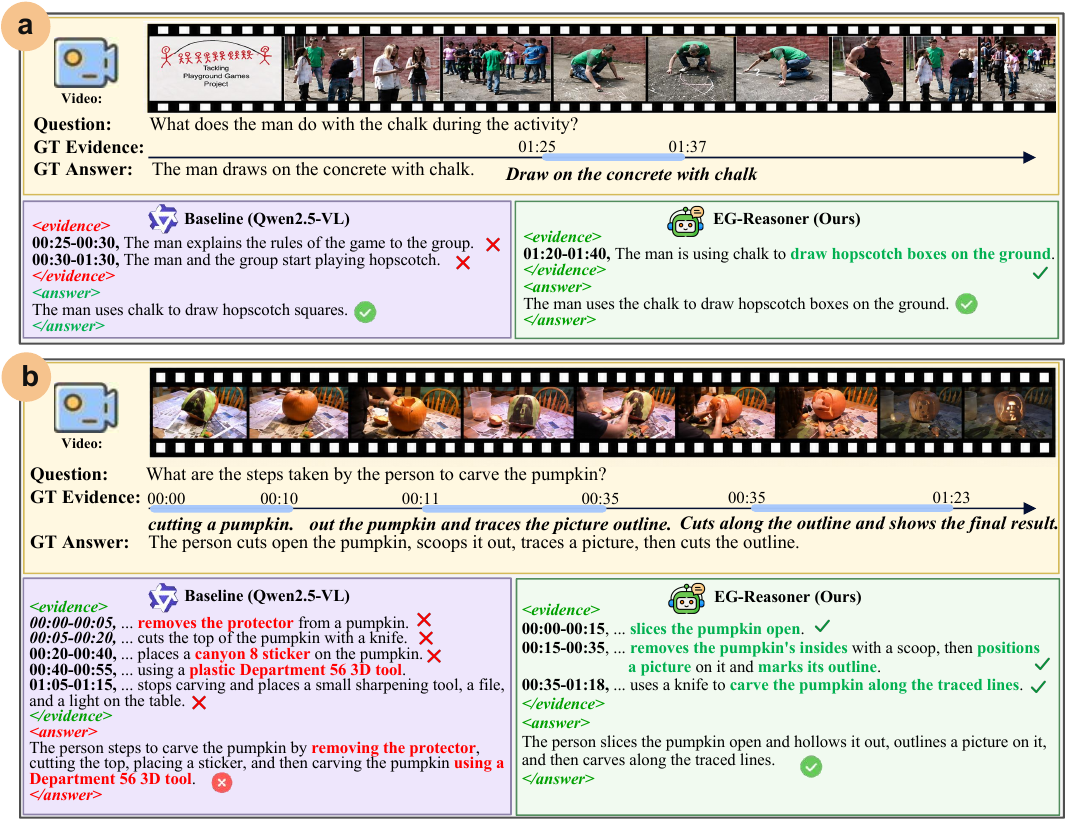}
    \vspace{-15pt}
    \caption{Qualitative comparison between the baseline model and EG-Reasoner. Hallucinated evidence is highlighted in red, and correctly grounded evidence is marked in green. }
    % \vspace{-5pt}
    \label{fig:case_study}
\end{figure}

\vspace{-5pt}
Figure~\ref{fig:case_study} presents qualitative comparisons between the base model Qwen2.5-VL and EG-Reasoner in evidence-grounded reasoning. In Figure~\ref{fig:case_study}(a), the baseline model predicts the correct answer but fails to localize the relevant temporal segment, instead relying on loosely related context. In contrast, EG-Reasoner accurately identifies the key evidence region associated with the chalk-drawing action. Figure~\ref{fig:case_study}(b) further demonstrates the advantage of evidence-grounded reasoning in complex procedural understanding. The baseline model produces hallucinated objects and incorrect reasoning steps, whereas EG-Reasoner decomposes the task into fine-grained temporal evidence segments and generates a coherent reasoning chain aligned with the video content, leading to the correct answer. Additional qualitative examples and failure cases are provided in Appendix~\ref{app:additional_cases}.
%---------------------------------------------------------------------------

%-----------------------------------6 Conclusion----------------------------
%---------------------------------------------------------------------------
\vspace{-5pt}
\section{Conclusion}
\vspace{-5pt}
This paper introduces EG-VQA, a benchmark that evaluates VideoQA models through joint assessment of answer correctness and temporal-semantic evidence faithfulness. Additionally, the Evidence-Grounded F1 (EG-F1) metric is introduced to jointly measure temporal alignment and semantic consistency between predicted and reference evidence. Experiments show that Video-LLMs, including strong proprietary models, may achieve high answer correctness while still exhibiting gaps in temporal-semantic evidence grounding, highlighting the discrepancy between answer correctness and evidence faithfulness. To further investigate evidence-aware learning under EG-VQA, we develop EG-Reasoner as an evidence-aware reference model trained with explicit evidence supervision. EG-Reasoner achieves the strongest evidence grounding performance among open-source models while maintaining competitive answer generation performance. Overall, our findings suggest that evidence-aware supervision provides an effective direction for improving grounded video understanding, and we hope EG-VQA can serve as a useful testbed for future research on evidence-aware multimodal reasoning.

\clearpage
% \section*{AI Use Statement}
% \vspace{-5pt}
% In this work, we used generative AI tools for both research-related tasks and manuscript preparation. For dataset construction, GPT-4o was used to generate candidate question--answer pairs and supporting evidence. DeepSeek-V3.2 and Gemini-2.5-Flash were used as independent evaluators during automatic quality control, with model disagreements further reviewed by human annotators. Gemini-2.5-Pro was used as an automatic judge for open-ended answer evaluation, and its agreement with human annotations was separately validated as described in the paper.

% Generative AI tools were also used to assist in drafting portions of the manuscript and to improve the readability and language quality of the paper. All AI-assisted text was reviewed, revised, and verified by the authors.

% The authors reviewed all AI-assisted research artifacts and manuscript content and take responsibility for the final content of this work, including all text, claims, experimental results, and conclusions.

%----------------------------------- Reference -----------------------------
%---------------------------------------------------------------------------
% \clearpage
\bibliography{iclr2027_conference}
\bibliographystyle{iclr2027_conference}

%----------------------------------- Appendix ------------------------------
%---------------------------------------------------------------------------
\clearpage
\appendix
\section{Dataset Construction Details}
\vspace{-0.5em}
\label{app:construction}

\subsection{Filtering protocol}
\vspace{-0.5em}
\label{app:filtering}

To ensure the reliability and consistency of the benchmark, we apply a structured filtering protocol to the collected videos and their associated annotations. This process is designed to correct annotation errors, remove low-quality samples, and retain videos that support meaningful reasoning.

\paragraph{Unified Metadata Format.}
The source datasets used in EG-VQA adopt different annotation schemas and temporal segmentation formats. To facilitate unified processing and evidence-grounded question generation, all videos are first converted into a standardized metadata representation consisting of video-level metadata, temporally ordered timestamps, and corresponding event descriptions.

Each video is represented as a structured sequence of timestamp--description pairs. An example is shown below:

\begin{verbatim}
  "TEOPA76qOKA": {
    "duration": 88,
    "title": "How to Make a Lemon Battery",
    "timestamps": [
      [30.0, 34.0],
      [34.0, 53.0],
      [53.0, 63.0],
      [63.0, 76.0],
      [76.0, 85.0]
    ],
    "descriptions": [
      "Roll the lemons",
      "Put copper wire and paper clips",
      "Connect alligator clips",
      "Touch the unconnected ends",
      "Connect the positive to LED"
    ]
  }
\end{verbatim}

This unified representation serves as the basis for subsequent filtering, question generation, evidence annotation, and quality control procedures.

\paragraph{Data Validation.}
We first remove corrupted, incomplete, or inaccessible video samples. Videos that cannot be decoded, contain missing segments, or fail to load correctly are excluded from further processing.

\paragraph{Temporal Alignment Refinement.}
Although the source datasets provide temporally segmented annotations, the original segment boundaries and descriptions may not always accurately reflect the underlying video content. To address this, human annotators are instructed to review each video and its corresponding segments, and to refine temporal boundaries when necessary. Segments that exhibit significant misalignment with the actual visual events are corrected or removed.

\paragraph{Description Consistency Check.}
We further examine the consistency between segment descriptions and the observed video content. Descriptions that are ambiguous, incomplete, or inconsistent with the visual evidence are revised or discarded. Samples with severely noisy or unreliable metadata are excluded entirely.

\paragraph{Informational Density Filtering.}
To ensure that the resulting dataset supports non-trivial reasoning, we remove videos with low informational density. Specifically, videos that contain only a single event, minimal temporal variation, or insufficient interaction between entities are filtered out, as they tend to yield trivial or purely descriptive questions.

\paragraph{Human Verification.}
All refinement steps are conducted by trained human annotators following a standardized review protocol. Annotators are instructed to prioritize both temporal accuracy and semantic consistency when validating segment annotations.

After applying the above filtering criteria, we retain a total of 2,067 high-quality videos, which serve as the foundation for constructing the EG-VQA benchmark.

% \vspace{-0.5em}
\subsection{Prompt Design and Iterative Refinement}
\vspace{-0.5em}
\label{app:prompt_design}

We construct question–answer (QA) pairs using a human-in-the-loop pipeline that combines large language model generation with iterative prompt refinement. The goal is to produce high-quality, reasoning-oriented questions that are grounded in temporally annotated video content.

\paragraph{Initial Prompt Design.}
Given the temporally segmented video metadata, we design type-specific prompts for each reasoning category (descriptive, temporal, causal, and counterfactual). Each prompt explicitly encourages the generation of open-ended questions that require reasoning over one or multiple temporal segments, rather than relying on superficial patterns or static descriptions.

\paragraph{Failure Mode Identification.}
Direct application of large language models often results in suboptimal QA pairs. To systematically identify common issues, we apply the initial prompts to a held-out subset of videos and manually inspect the outputs. The most frequent failure modes include:
\begin{itemize}
    \item \textbf{Video-independent questions:} Questions that can be answered without referring to the video content.
    \item \textbf{Weak grounding:} Questions or answers that are only loosely related to the temporal metadata.
    \item \textbf{Incomplete reasoning:} Answers that omit necessary temporal or contextual information.
\end{itemize}

\paragraph{Iterative Refinement.}
Based on these observations, we iteratively refine the prompts by introducing stricter constraints on grounding, temporal specificity, and reasoning depth. This process follows a generate–inspect–refine loop, where prompts are repeatedly adjusted and re-evaluated until the generated QA pairs consistently satisfy our quality criteria. During this process, we explicitly filter templated or weakly grounded questions to reduce potential shortcut patterns that could be exploited without genuine video understanding.

\paragraph{Final Deployment.}
Once stabilized, the refined prompts are applied to the full set of filtered videos to generate candidate QA pairs at scale. This iterative design ensures that the resulting dataset emphasizes non-trivial, reasoning-intensive, and evidence-grounded questions.

% \vspace{-0.5em}
\subsection{Quality Control Protocol}
\vspace{-0.5em}
\label{app:quality_control}

To ensure the reliability of generated QA pairs, we employ a two-stage quality control protocol that combines structured automatic evaluation with human verification.

\paragraph{Stage 1: Structured Multi-model Evaluation.}
Each candidate QA pair is independently evaluated by two large language models from distinct architectural families (DeepSeek-V3.2 and Gemini-2.5-Flash). We adopt a structured scoring rubric consisting of three criteria:
\begin{itemize}
    \item \textbf{Question Reasonableness:} whether the question is well-posed and requires video understanding.
    \item \textbf{Answer Correctness:} whether the answer is consistent with the video content.
    \item \textbf{Evidence Alignment:} whether the answer is properly grounded in the relevant temporal segments.
\end{itemize}

Each dimension is scored on a scale of 0–10, and the total score is obtained by summing the three dimensions. A sample is considered acceptable if the total score is greater than or equal to 25.

\paragraph{Agreement-based Filtering.}
We adopt a dual-model agreement strategy to improve robustness. When both models agree on the decision (accept or reject), the result is directly adopted. Otherwise, the sample is flagged for manual review.

\paragraph{Stage 2: Human Adjudication.}
All samples for which the two evaluators disagree are reviewed by human annotators using the same criteria of question reasonableness, answer correctness, and evidence alignment. Annotators determine whether a sample should be retained, corrected, or removed.

\paragraph{Post-hoc Validation.}
% 原版
% After the full-dataset screening and adjudication process, we conduct an additional stratified random audit of 1,000 QA pairs. The audit is stratified across the three source datasets, video-duration ranges, and four question types to provide broad coverage of the resulting benchmark. More than 95\% of the audited samples are verified as valid.  
After the full-dataset screening and adjudication process, we conduct an additional stratified random audit of 1,000 QA pairs. The audit is stratified across the three source datasets, video-duration ranges, and four question types to provide broad coverage of the resulting benchmark. For each audited sample, human annotators inspect the original video together with the corresponding metadata, generated question, reference answer, and supporting evidence. They verify whether the question is answerable from the observed video content, whether the answer is correctly supported by the video, and whether the annotated evidence is temporally and semantically aligned with the answer. Of the 1,000 audited samples, 932 (93.2\%) are judged valid without modification, while 24 (2.4\%) contain only minor correctable issues and are retained after manual revision.

\paragraph{Invalid-sample Handling.}
% 原版
% Audited samples with minor and correctable issues, such as small inconsistencies in wording or evidence annotation, are manually revised and retained. Samples with substantive problems, including unanswerable questions, incorrect answers, or severe answer–evidence inconsistencies, are removed.  
The remaining 44 samples (4.4\%) are removed due to substantive validity issues, including questions that are not answerable from the video, answers that are inconsistent with the observed content, supporting evidence that fails to justify the answer. Overall, 956 of the 1,000 audited samples (95.6\%) are retained after post-hoc validation and correction.

% \paragraph{Failure-pattern Follow-up.}
% We further inspect the dataset for the failure patterns observed during the audit. This follow-up inspection does not reveal additional systematic error modes, suggesting that the remaining residual noise after the multi-stage quality-control process is limited.  

% \paragraph{Inter-annotator Consistency.}
% Human disagreements are resolved through discussion, and the annotation guidelines are refined when necessary to maintain consistent quality-control decisions.  

\vspace{-0.5em}
\subsection{Dataset Statistics and Benchmark Comparison}
% \vspace{-0.5em}
\label{app:statistics}

\begin{figure}[h]
    \centering
    \includegraphics[width=\linewidth]{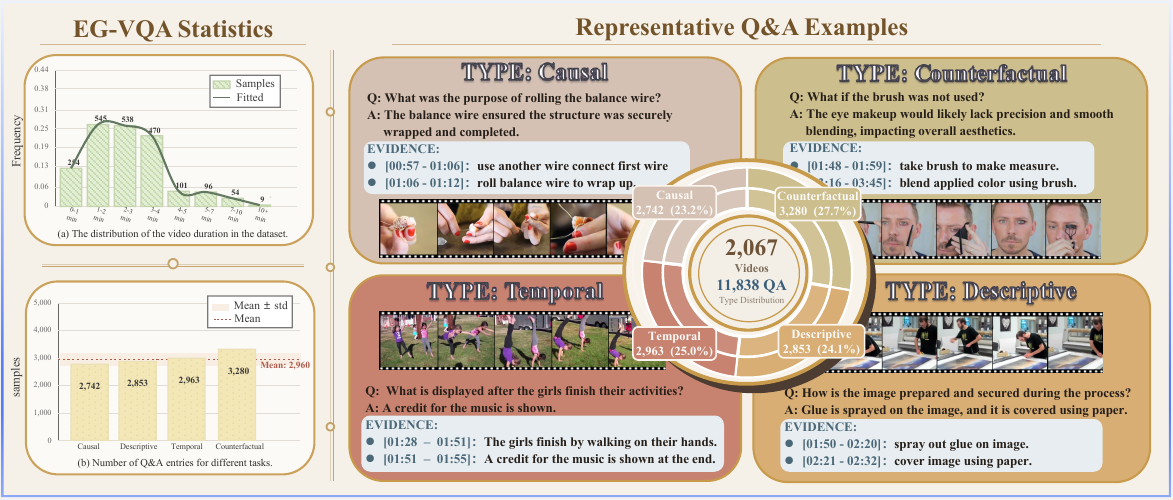}
    \vspace{-1.5em}
    \caption{
    Overview of EG-VQA statistics and representative QA examples.
    }
    \label{fig:egvqa_statistics}
    \vspace{-6pt}
\end{figure}

% \vspace{-5pt}

As shown in Figure~\ref{fig:egvqa_statistics}, the EG-VQA benchmark contains 2,067 videos and 11,838 QA pairs. The benchmark is constructed from the official validation splits of ActivityNet Captions and YouCook2, and the validation and test splits of HiREST. After source-specific curation, we perform a stratified video-level split by controlling the distributions of source dataset, video duration, and evidence density. All QA pairs and evidence annotations derived from the same video are assigned to the same partition, ensuring that no video appears in both the training and test sets. Detailed source-wise statistics are summarized in Table~\ref{tab:source_split_stats}, while per-split question statistics are provided in Table~\ref{tab:split_stats}.

\paragraph{Reasoning-oriented Question Distribution.}
EG-VQA contains four categories of reasoning tasks: descriptive, temporal, causal, and counterfactual. These categories progressively increase the reasoning difficulty, ranging from direct perception of observable content to hypothetical reasoning over altered events. As shown in Figure~\ref{fig:egvqa_statistics}, the dataset maintains a relatively balanced distribution across different reasoning types, ensuring diverse evaluation of video understanding capability.

% \paragraph{Evidence-grounded Reasoning Characteristics.}
% EG-VQA explicitly associates each question with one or more temporally localized evidence units. Each evidence unit consists of a temporal span and a textual description of the corresponding event, enabling models to identify both where supporting evidence occurs and what semantic content it conveys. On average, each question is paired with 2.27 evidence segments, reflecting the multi-step nature of video reasoning.

\begin{table}[h]
\centering
\caption{Source composition and video-level split statistics of EG-VQA.}
\vspace{-10pt}
\label{tab:source_split_stats}
\small
\setlength{\tabcolsep}{12pt}
\renewcommand{\arraystretch}{1.15}
\begin{tabular}{lcccc}
\toprule
Source Dataset & Source Split & \#Videos & Train & Test \\
\midrule
ActivityNet Captions & Validation & 1,500 & 1,277 & 223 \\
YouCook2 & Validation & 209 & 168 & 41 \\
HiREST & Val. + Test & 358 & 286 & 72 \\
\midrule
Total & -- & 2,067 & 1,731 & 336 \\
\bottomrule
\end{tabular}
\end{table}

\vspace{-5pt}

\begin{table}[h]
\centering
\caption{Question and evidence statistics of EG-VQA.}
\vspace{-10pt}
\label{tab:split_stats}
\small
\setlength{\tabcolsep}{12pt}
\renewcommand{\arraystretch}{1.15}
\begin{tabular}{lcccc}
\toprule
& \multicolumn{2}{c}{Train} & \multicolumn{2}{c}{Test} \\
\cmidrule(lr){2-3}
\cmidrule(lr){4-5}
Question Type & \#Pairs & Avg. Evid. &
\#Pairs & Avg. Evid. \\
\midrule
Descriptive & 2,196 & 2.12 & 657 & 2.24 \\
Temporal & 2,212 & 2.66 & 751 & 2.89 \\
Causal & 1,994 & 2.23 & 748 & 2.26 \\
Counterfactual & 2,547 & 1.96 & 733 & 2.04 \\
\midrule
Total & 8,949 & 2.24 & 2,889 & 2.36 \\
\bottomrule
\end{tabular}
\end{table}
% \vspace{-5pt}

\paragraph{Evidence-grounded Reasoning Characteristics.}
EG-VQA explicitly associates each question with one or more temporally localized evidence units. Each evidence unit consists of a temporal span and a textual description of the corresponding event, enabling models to identify both where supporting evidence occurs and what semantic content it conveys. On average, each question is paired with 2.27 evidence segments, reflecting the multi-step nature of video reasoning.

\begin{figure}[ht]
    \centering
    \includegraphics[width=1.0\linewidth]{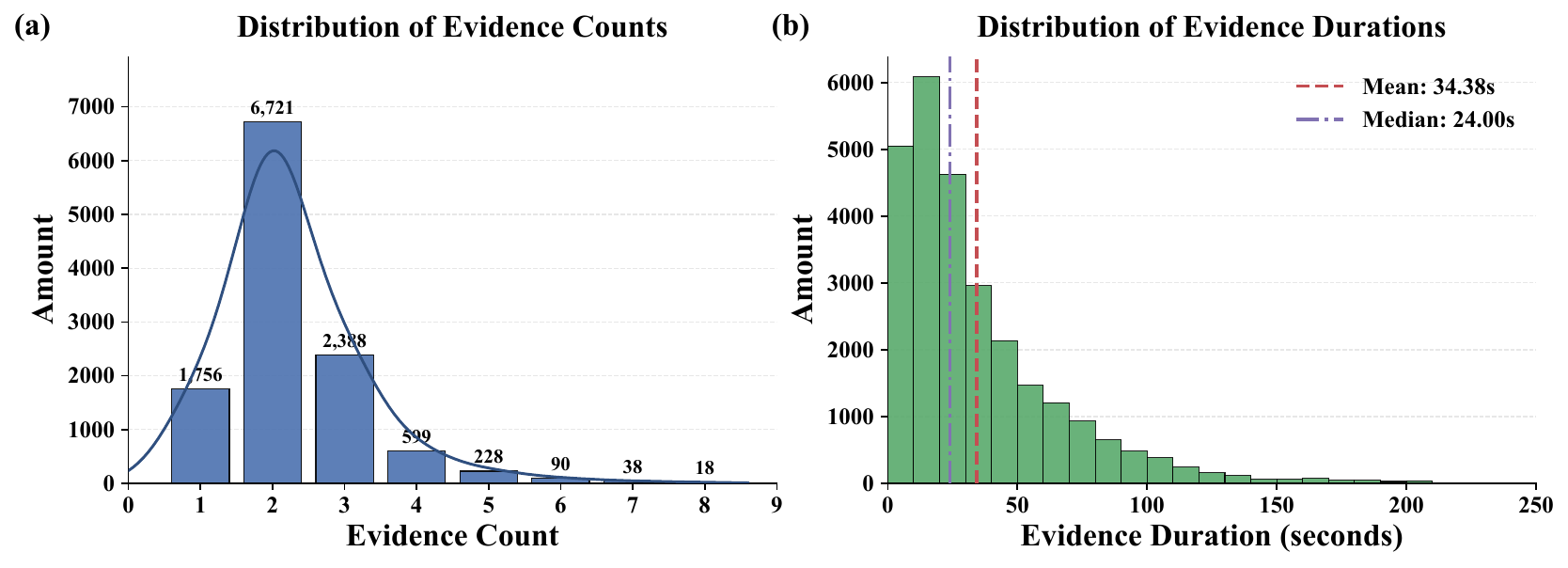}
    \vspace{-1.0em}
    \caption{
    Evidence statistics of EG-VQA.
    (a) Distribution of evidence counts per question.
    (b) Distribution of evidence duration.
    }
    \label{fig:evidence_count&duration_distribution}
    \vspace{-6pt}
\end{figure}

Figure~\ref{fig:evidence_count&duration_distribution}(a) shows the distribution of evidence counts per question. Most questions require multiple evidence segments rather than a single localized event, indicating that successful reasoning often depends on integrating information across temporally separated video events.

In addition, the temporal spans of evidence vary substantially across samples. Figure~\ref{fig:evidence_count&duration_distribution}(b) illustrates the distribution of evidence duration. The benchmark contains both short fine-grained actions and long procedural activities, requiring models to perform reasoning at multiple temporal scales.

The average video duration is 177.6 seconds, while the average evidence segment length is 34.4 seconds. This large gap further increases the difficulty of evidence grounding, since models must identify relevant moments within long video contexts.

\paragraph{Benchmark Comparison.}
Table~\ref{tab:benchmark_comparison} compares EG-VQA with representative VideoQA benchmarks that provide explicit evidence-related supervision. Existing grounded benchmarks have made important progress in associating answers with relevant temporal evidence. However, they typically represent evidence primarily through temporal localization annotations. In contrast, EG-VQA explicitly represents supporting evidence as temporal-semantic units, where each evidence segment is paired with a textual description of the corresponding event.

% \vspace{-8pt}
\begin{table}[ht]
\centering
\caption{Comparison between evidence-grounded VideoQA benchmarks and EG-VQA.}
\vspace{-10pt}
\label{tab:benchmark_comparison}
\small
\setlength{\tabcolsep}{2.0pt}
\renewcommand{\arraystretch}{1.15}

\begin{tabular}{lccccc}
\toprule
\textbf{Benchmark} &
\textbf{Format} &
\textbf{Temporal} &
\textbf{Multiple} &
\textbf{Textual Evidence} &
\textbf{Semantic Matching} \\
\midrule

TVQA+~\citep{lei2020tvqa+}
& MC 
& $\checkmark$ 
& $\times$ 
& $\times$ 
& $\times$ \\

NExT-GQA~\citep{xiao2024can} 
& MC 
& $\checkmark$ 
& $\checkmark$ 
& $\times$ 
& $\times$ \\

MultiHop-EgoQA~\citep{Chen2024GroundedMV}
& OE 
& $\checkmark$ 
& $\checkmark$ 
& $\times$ 
& $\times$ \\

% TOGA~\citep{Gupta2025TOGATG}
% & OE 
% & $\checkmark$ 
% & $\times$ 
% & $\times$ 
% & $\times$ \\

CG-Bench~\citep{Chen2024CGBenchCQ}
& MC 
& $\checkmark$ 
& $\checkmark$ 
& $\times$ 
& $\times$ \\

HERBench~\citep{ben2026herbench}
& MC 
& $\times$ 
& $\checkmark$ 
& $\times$ 
& $\times$ \\

\midrule

\textbf{EG-VQA} 
& \textbf{OE} 
& \textbf{$\checkmark$} 
& \textbf{$\checkmark$} 
& \textbf{$\checkmark$} 
& \textbf{$\checkmark$} \\

\bottomrule
\end{tabular}

\vspace{1mm}

{\footnotesize
\justifying
\textit{Note.}
Here, ``Temporal'' denotes temporal evidence localization,
``Multiple'' denotes multi-segment evidence annotation,
``Textual Evidence'' denotes standalone textual descriptions for evidence segments,
and ``Semantic Matching'' denotes semantic evaluation of evidence descriptions.\par
}
\end{table}
\vspace{-5pt}

Therefore, EG-VQA enables evaluation beyond whether a model identifies relevant moments, by additionally assessing whether the generated evidence descriptions are semantically consistent with the supporting video events. This design supports research on open-ended VideoQA with explicitly verifiable temporal-semantic evidence.

\section{EG-F1: Complete Formulation}
\label{app:egf1_details}

This section expands the EG-F1 metric introduced in Section~\ref{sec:egf1}.

\subsection{Computation Steps}
\paragraph{Step 1: Pairwise similarity.} 
For a ground-truth item $g_i=(t_i,d_i)$ and a prediction $p_j=(t_j,d_j)$, we compute both temporal overlap and semantic similarity:
\begin{align}
    \mathrm{IoU}_{ij} &= \frac{|t_i \cap t_j|}{|t_i \cup t_j|}, \\
    \mathrm{Sim}_{ij} &= \cos\!\bigl(\phi(d_i), \phi(d_j)\bigr),
\end{align}
where $\phi(\cdot)$ denotes a sentence embedding function used to encode textual evidence descriptions.

\paragraph{Step 2: Edge construction.} Valid candidate edges are defined by
\begin{equation}
    \mathcal{E} = \{(i,j) \mid \mathrm{IoU}_{ij} \geq \alpha \land \mathrm{Sim}_{ij} \geq \beta\},
\end{equation}
where $\alpha$ and $\beta$ denote the temporal IoU threshold and semantic similarity threshold, respectively. Different threshold settings are used in our experiments to evaluate grounding performance under varying levels of strictness.

\paragraph{Step 3: Optimal matching.} A bipartite graph with edge weights
\begin{equation}
    s_{ij} = 
    \begin{cases}
        \mathrm{IoU}_{ij} \cdot \mathrm{Sim}_{ij}, & (i,j) \in \mathcal{E}, \\
        0, & \text{otherwise},
    \end{cases}
\end{equation}
is constructed between $\mathcal{G}$ and $\mathcal{P}$. The Hungarian algorithm is applied to obtain a maximum-weight one-to-one matching $\mathcal{M}$. Let $M = |\{(i,j)\in\mathcal{M} \mid s_{ij}>0\}|$.

\paragraph{Step 4: Metric computation.} Precision, recall, and F1 are computed as
\begin{align}
    P &= \frac{M}{|\mathcal{P}|}, \quad R = \frac{M}{|\mathcal{G}|}
\end{align}
\begin{align}
    \mathrm{EG\!-\!F1} = \frac{2PR}{P+R}.
\end{align}

\subsection{Soft variant for training.}
\vspace{-0.5em}
\label{app:soft_variant}
To mitigate reward sparsity during reinforcement learning, a soft variant is employed by removing the hard threshold in Step 2. The pairwise weight becomes $s_{ij}^{\mathrm{soft}} = \mathrm{IoU}_{ij} \cdot \mathrm{Sim}_{ij}$ for all $(i,j)$, and soft precision/recall are defined as
\begin{equation}
    P_{\mathrm{soft}} = \frac{\sum_{(i,j)\in\mathcal{M}} s_{ij}^{\mathrm{soft}}}{|\mathcal{P}|}, \quad
    R_{\mathrm{soft}} = \frac{\sum_{(i,j)\in\mathcal{M}} s_{ij}^{\mathrm{soft}}}{|\mathcal{G}|}.
\end{equation}
The soft variant is then 
\begin{equation}
\text{EG-F1}_{\mathrm{soft}} = \frac{2 P_{\mathrm{soft}} R_{\mathrm{soft}}}{P_{\mathrm{soft}} + R_{\mathrm{soft}}}
\end{equation}
, which is combined with the hard EG-F1 as described in the main text.

\subsection{Complexity.} Computing all pairwise similarities costs $O(|\mathcal{G}||\mathcal{P}|)$; the Hungarian algorithm adds $O(\max(|\mathcal{G}|,|\mathcal{P}|)^3)$. In practice, $|\mathcal{G}|$ and $|\mathcal{P}|$ are small (average 2.27), making the metric efficient to compute at both training and inference time.

\section{EG-Reasoner: Supplementary Details}
\label{app:training}

\subsection{Answer Reward Details}
\label{app:answer_reward}

The answer reward evaluates whether the generated answer is semantically correct and consistent with the reference answer while avoiding hallucinated or unsupported content.

\paragraph{Evaluation Principles.}
% We adopt Gemini-2.5-Flash as a rule-guided LLM evaluator that assesses each generated answer based on two complementary principles:
We adopt a rule-guided LLM evaluator that assesses each generated answer based on two complementary principles:
\begin{itemize}
    \item \textbf{Inclusion principle:} the generated answer should cover as many key elements of the reference answer as possible.
    \item \textbf{Exclusion principle:} the answer should not contain information that contradicts or cannot be inferred from the video metadata.
\end{itemize}

\paragraph{Scoring Scheme.}
To reduce reward sparsity, especially during early training, we employ a relaxed scoring scheme that assigns partial credit:
\begin{equation}
R_{\mathrm{ans}}(y) =
\begin{cases}
1, & \text{fully correct}, \\
0.5, & \text{partially correct}, \\
0, & \text{incorrect}.
\end{cases}
\end{equation}

This design provides denser supervision signals compared to binary evaluation, enabling more stable optimization.

\paragraph{Prompt-based Evaluation.}
The evaluation is performed using a prompt-based LLM judge, which compares the generated answer against the reference answer and the associated video metadata. The prompt explicitly instructs the evaluator to follow the inclusion and exclusion principles and to output a discrete score according to the predefined rubric.

\subsection{GRPO Objective Details}
\label{app:grpo}

We adopt Group Relative Policy Optimization (GRPO) to optimize the structured generation policy.

Given an input $(V, q)$, the policy $\pi_\theta$ samples a group of candidate responses $\{y_1, \dots, y_G\}$. Each candidate is assigned a reward $R_i$, and the relative advantage is computed as:
\begin{equation}
A_i = \frac{R_i - \mu}{\sigma},
\end{equation}
where $\mu$ and $\sigma$ denote the mean and standard deviation of rewards within the sampled group.

Let $r_i = \frac{\pi_\theta(y_i \mid V,q)}{\pi_{\theta_{\mathrm{old}}}(y_i \mid V,q)}$ denote the importance sampling ratio. The clipped GRPO objective is:
\begin{equation}
\mathcal{L}(\theta) = 
\mathbb{E}_{(V,q), \{y_i\} \sim \pi_{\theta_{\mathrm{old}}}}
\left[
\sum_{i=1}^G 
\min\!\left(
r_i A_i,\,
\operatorname{clip}(r_i, 1-\epsilon, 1+\epsilon) A_i
\right)
- \beta_{\mathrm{KL}} \, \mathrm{KL}(\pi_\theta \| \pi_{\mathrm{ref}})
\right],
\end{equation}
where $\epsilon=0.2$ is the clipping parameter and $\beta_{\mathrm{KL}}=0.01$ controls the strength of KL regularization.

\paragraph{KL Regularization.}
The KL term discourages excessive deviation from the reference policy during post-training. It is evaluated over the generated tokens and added to the policy objective as a regularization term.

GRPO uses group-relative rewards to estimate advantages and therefore does not require a separately trained value model. We otherwise follow the standard GRPO optimization procedure.

\subsection{Training Configuration}
\label{app:hyperparams}
The training configuration of EG-Reasoner is summarized in Table~\ref{tab:training_config}. We initialize the policy from Qwen2.5-VL-7B-Instruct, freeze the visual backbone, and optimize only the language-model parameters. Unless otherwise specified, all EG-Reasoner variants use the same training configuration.

\vspace{-5pt}
\begin{table}[!ht]
\centering
\small
\caption{Training configuration of EG-Reasoner.}
\vspace{-10pt}
\renewcommand{\arraystretch}{1.15}
\begin{tabular}{lc}
\toprule
\textbf{Hyperparameter} & \textbf{Value} \\
\midrule
Base model & Qwen2.5-VL-7B-Instruct \\
Trainable parameters & Language model only \\
Frozen parameters & Visual backbone \\
Epochs & 1 \\
Training frames & 32 \\
Frame sampling strategy & Uniform sampling \\
Per-device batch size & 4 \\
Learning rate & 1e-6 \\
Training steps & 558 \\
Group size $G$ & 8 \\
$\lambda_f$ / $\lambda_e$ / $\lambda_a$ & 0.1 / 0.6 / 0.3 \\
$\epsilon$ (clip) & 0.2 \\
$\beta_{\mathrm{KL}}$ & 0.01 \\
Sentence Embeddings & all-MiniLM-L6-v2 \\
Number of GPUs & 4 $\times$ NVIDIA A800 \\
Training time & $\sim$50 hours \\
Output format & Unified structured format in Section~4 \\
\bottomrule
\end{tabular}

\label{tab:training_config}
\end{table}
\vspace{-5pt}

\section{Experimental Details}
\label{app:experimental_details}

\subsection{Baseline Details}
\label{app:baseline_details}

We evaluate representative Video-LLMs covering proprietary systems, general-purpose open-source models, and recent reasoning-enhanced variants.

\paragraph{GPT-4o~\citep{hurst2024gpt}}
is a proprietary multimodal large language model capable of processing visual and textual inputs in a unified framework. We use it as a representative commercial baseline.

\paragraph{Gemini-2.5-Flash~\citep{comanici2025gemini}}
is a proprietary multimodal model from the Gemini family, designed to balance efficiency and reasoning capability. We include it to evaluate commercial multimodal systems.

\paragraph{GPT-5.6 Luna}
is a proprietary multimodal model from OpenAI, designed for general-purpose multimodal understanding and reasoning. We include it to assess the capability of recent commercial models.

\paragraph{Qwen3.7 Plus}
is a proprietary multimodal model from the Qwen family, developed for multimodal perception and reasoning. We include it as a representative commercial model.

\paragraph{Qwen2.5-VL-7B~\citep{Qwen2.5-VL}}
is an open-source multimodal model built upon the Qwen2.5 backbone. It supports image and video inputs and is used as a representative general-purpose open-source baseline.

\paragraph{InternVL-3-8B~\citep{zhu2025internvl3}}
is an open-source multimodal model from the InternVL family, designed for multimodal perception and reasoning. It is included as a general multimodal understanding baseline.

\paragraph{Time-R1-7B~\citep{wang2026time}}
is a reasoning-oriented video model designed to improve temporal reasoning and event-level understanding. We include it to evaluate temporal reasoning capability.

\paragraph{TW-GRPO~\citep{dang2025reinforcing}}
is a GRPO-based reasoning model that introduces token-wise weighting and soft reward mechanisms to provide more fine-grained supervision. It is included as a reinforcement learning-based reasoning baseline.

\paragraph{VideoChat-R1-7B~\citep{li2025videochat}}
is a video reasoning model trained with reinforcement learning to enhance spatio-temporal perception and reasoning. It serves as a strong open-source video reasoning baseline.

\paragraph{VideoChat-R1-Thinking-7B~\citep{li2025videochat}}
extends the VideoChat-R1 framework by incorporating explicit thinking-style reasoning, and is included as a reasoning-oriented video baseline.

\paragraph{VideoChat-R1.5-7B~\citep{yan2026videochat}}
is an improved version of VideoChat-R1 that emphasizes test-time perceptual refinement. We include it as a recent video reasoning baseline.

\subsection{Answer Evaluation Protocol}
\label{app:answer_eval}

We evaluate answer correctness using an LLM-based protocol, as EG-VQA is formulated as open-ended question answering where semantically equivalent answers may have different surface forms, making traditional string-based metrics insufficient.

Given a test set of $N$ samples, let $x_i$ denote the model response and $y_i$ the reference answer. The strict accuracy is defined as:
\begin{equation}
\mathrm{Acc}_{\mathrm{strict}}
=
\frac{1}{N}
\sum_{i=1}^{N}
\mathrm{Match}_{\mathrm{full}}(x_i, y_i),
\end{equation}
where $\mathrm{Match}_{\mathrm{full}}(\cdot)$ returns 1 if the response fully matches the reference answer in meaning.

To account for partially correct answers, we define relaxed accuracy as:
\begin{equation}
\mathrm{Acc}_{\mathrm{relaxed}}
=
\frac{1}{N}
\sum_{i=1}^{N}
\left(
\mathrm{Match}_{\mathrm{full}}(x_i, y_i)
+
\gamma \cdot
\mathrm{Match}_{\mathrm{partial}}(x_i, y_i)
\right),
\end{equation}
where $\mathrm{Match}_{\mathrm{partial}}(\cdot)$ indicates partial correctness and $\gamma$ controls the credit assigned to partially correct answers. In our experiments, $\gamma=0.5$.

We use Gemini-2.5-Pro as the evaluator. For each prediction, the evaluator receives the question, reference answer, model response, video title, and temporally annotated video metadata, and assigns one of three labels: Fully Correct, Partially Correct, or Incorrect. The same protocol is applied to all evaluated models.

% 改动前
% To validate the reliability of the evaluator, we construct a set of 200 QA instances balanced across the three data sources and four question types. Annotator 1 constructs 200 candidate responses comprising 100 intended Fully Correct, 50 Partially Correct, and 50 Incorrect cases. Annotator 2, blind to these preset labels, independently evaluates all responses using the same $1/0.5/0$ scoring rubric. Four responses initially designated as Fully Correct are reclassified as Partially Correct, resulting in 96 Fully Correct, 54 Partially Correct, and 50 Incorrect human judgments. Gemini-2.5-Pro independently evaluates the same inputs under the same protocol.
To validate the reliability of the evaluator, we collect responses generated by Qwen2.5-VL-7B on the EG-VQA evaluation set. Annotator 1 manually inspected these real model outputs and selected a balanced validation subset of 200 responses, including 100 Fully Correct, 50 Partially Correct, and 50 Incorrect cases. The selected samples are balanced across the three source datasets and four question types. Annotator 2, blind to Annotator 1's selections, independently evaluated the same responses using the 1/0.5/0 scoring rubric. Four responses selected as Fully Correct by Annotator 1 were reclassified as Partially Correct by Annotator 2, resulting in 96 Fully Correct, 54 Partially Correct, and 50 Incorrect human judgments. Gemini-2.5-Pro independently evaluated the same inputs under the same protocol.

\vspace{-5pt} % 缩小上文与表格之间的间距

\begin{table}[h]
\centering
\caption{Confusion matrix between human and Gemini-2.5-Pro judgments.}
\vspace{-10pt}
\label{tab:judge_confusion}
\small
\setlength{\tabcolsep}{13pt}
\renewcommand{\arraystretch}{1.1}
\begin{tabular}{ccccc}
\toprule
\diagbox[width=13em,height=2.4em]{Human}{Gemini-2.5-Pro}
& Incorrect & Partial & Full & Total \\
\midrule
Incorrect & 50 & 0  & 0   & 50  \\
Partial   & 0  & 47 & 7   & 54  \\
Full      & 0  & 0  & 96  & 96  \\
\midrule
Total     & 50 & 47 & 103 & 200 \\
\bottomrule
\end{tabular}
\end{table}
\vspace{-5pt}

Gemini-2.5-Pro achieves 96.5\% exact agreement with human judgments (193/200), with a Cohen's $\kappa$ of 0.944. All seven disagreements occur between Partially Correct and Fully Correct judgments, with none crossing the Incorrect-versus-Correct boundary. In particular, all 50 responses judged Incorrect by human annotators are also identified as Incorrect by Gemini-2.5-Pro, indicating that the remaining discrepancies are confined to the boundary of answer completeness rather than substantive correctness.

\subsection{Evidence Evaluation Settings}
\label{app:evidence_eval}
We evaluate evidence grounding using both event-level temporal F1 and the proposed EG-F1 metric.

For temporal localization, we report F1 scores under multiple IoU thresholds $\tau \in \{0.1, 0.3, 0.5, 0.7\}$ to evaluate localization performance at different granularities.

For EG-F1, we evaluate multiple threshold combinations $(\alpha, \beta)$, where $\alpha$ denotes the temporal IoU threshold and $\beta$ denotes the semantic similarity threshold. Specifically, we report results under $(0.3, 0.5)$, $(0.3, 0.75)$, and $(0.5, 0.75)$, corresponding to increasingly strict grounding criteria.

\subsection{Grounded Answer Score}
\label{app:gas}

Answer correctness and evidence grounding measure complementary aspects of evidence-grounded VideoQA, but reporting them independently does not indicate whether both are achieved on the same samples. We therefore additionally report the Grounded Answer Score (GAS), an auxiliary sample-level metric that couples answer correctness with evidence quality.

For sample $i$, let $a_i \in \{0, 0.5, 1\}$ denote the relaxed answer score corresponding to Incorrect, Partially Correct, and Fully Correct predictions, respectively. Let $s_i^{\mathrm{soft}} \in [0,1]$ denote the Soft EG-F1 score of the predicted evidence, computed using the soft temporal-semantic matching formulation in Appendix~B.2. GAS is defined as:
\begin{equation}
\mathrm{GAS}
=
\frac{100}{N}
\sum_{i=1}^{N}
a_i \, s_i^{\mathrm{soft}}.
\end{equation}

A high GAS therefore requires a model to produce both a correct answer and well-grounded evidence for the same sample. Incorrect answers contribute zero regardless of evidence quality, while partially correct answers receive half credit. We use Soft EG-F1 as the evidence component to retain continuous temporal-semantic evidence quality without introducing an additional hard threshold. GAS is computed by coupling answer correctness and evidence quality at the sample level before dataset-level aggregation, and therefore differs from multiplying their independently averaged scores.
%---------------------------------------------------------------------------
\subsection{Evaluation Configuration}
\label{app:experimental_setup}

The key evaluation settings for our experiments on the EG-VQA benchmark are summarized in Table~\ref{tab:evaluation_config}.

% \vspace{-5pt} % 缩小上文与表格之间的间距

\begin{table}[ht]
\centering
\small
\caption{Evaluation configuration used for EG-VQA.}
\vspace{-10pt}
\renewcommand{\arraystretch}{1.15}
\begin{tabular}{lc}
\toprule
\textbf{Evaluation setting} & \textbf{Value} \\
\midrule
Benchmark & EG-VQA \\
Task format & Open-ended QA \\
Evaluation frames & 64 \\
Frame sampling strategy & Uniform sampling \\
Maximum visual resolution budget & $256 \times 28 \times 28$ \\
Output format & Unified structured answer-and-evidence format \\
Answer metric & Strict / Relaxed accuracy \\
Evidence metric & EG-F1 and event-level F1 \\
\bottomrule
\end{tabular}

\label{tab:evaluation_config}
\end{table}
%---------------------------------------------------------------------------
\subsection{Full Evidence Grounding Results}
\label{app:full_evidence_results}

Table~\ref{tab:full_evidence_results} reports the complete evidence grounding results under the evaluation settings defined in Appendix~\ref{app:evidence_eval}. These results complement the evidence-grounding summary reported in Table~\ref{tab:merged-experimental-evidence-results} with threshold-specific breakdowns.

\begin{table}[ht]
\begin{center}
\begin{minipage}[t]{\linewidth}
\centering
% \vspace{-5pt}
\captionof{table}{Threshold-specific evidence grounding results on EG-VQA.}
\vspace{-10pt}
\label{tab:full_evidence_results}
{\setlength{\tabcolsep}{4.0pt}
\renewcommand{\arraystretch}{1.08}
\resizebox{\linewidth}{!}{%
\begin{tabular}{l*{4}{c}*{3}{c}}
\toprule
\multirow{2}{*}{\textbf{Model}}
& \multicolumn{4}{c}{\textbf{Event-level F1}}
& \multicolumn{3}{c}{\textbf{EG-F1}} \\
\cmidrule(lr){2-5}
\cmidrule(lr){6-8}
& \textbf{@0.1}
& \textbf{@0.3}
& \textbf{@0.5}
& \textbf{@0.7}
& \shortstack{\textbf{(0.3, 0.5)}}
& \shortstack{\textbf{(0.3, 0.75)}}
& \shortstack{\textbf{(0.5, 0.75)}} \\
\midrule

\textbf{Human Results*}
& 93.29 & 89.48 & 68.75 & 43.52
& 84.61 & 47.67 & 36.74 \\

\midrule
\rowcolor{gray!15}
\multicolumn{8}{c}{\textit{Commercial Large Multimodal Models}} \\
\midrule

GPT-4o~\cite{hurst2024gpt}
& 35.85 & 17.85 & 5.67 & 2.95
& 8.49 & 1.18 & 0.33 \\

Gemini-2.5-Flash~\cite{comanici2025gemini}
& 28.87 & 19.81 & 12.56 & 6.88
& 12.64 & 1.48 & 0.86 \\

GPT-5.6 Luna
& 57.13 & 41.88 & 22.39 & 9.78
& 25.60 & 4.09 & 1.99 \\

Qwen3.7 Plus
& 36.64 & 25.18 & 14.63 & 6.49
& 15.57 & 1.37 & 0.73 \\

\midrule
\rowcolor{gray!15}
\multicolumn{8}{c}{\textit{Open-sourced Video-LLMs}} \\
\midrule

Qwen2.5-VL-7B~\cite{Qwen2.5-VL}
& 29.36 & 15.83 & 7.70 & 2.55
& 8.18 & 1.15 & 0.66 \\

InternVL-3-8B~\cite{zhu2025internvl3}
& 20.89 & 11.02 & 4.68 & 1.34
& 5.11 & 0.60 & 0.35 \\

Time-R1-7B~\cite{wang2026time}
& 46.46 & 32.03 & 15.64 & 5.41
& 16.46 & 2.75 & 1.40 \\

VideoChat-R1-7B~\cite{li2025videochat}
& 48.24 & 34.50 & 18.43 & 6.84
& 18.71 & 2.70 & 1.56 \\

VideoChat-R1-Thinking-7B~\cite{li2025videochat}
& 40.90 & 26.90 & 13.69 & 4.77
& 14.51 & 1.82 & 0.94 \\

VideoChat-R1.5-7B~\cite{yan2026videochat}
& 33.26 & 21.44 & 10.88 & 3.58
& 11.86 & 2.17 & 1.22 \\

TW-GRPO~\cite{dang2025reinforcing}
& 33.30 & 19.48 & 8.97 & 2.94
& 9.65 & 1.07 & 0.58 \\

\midrule

\textbf{EG-Reasoner (Ours)}
& \textbf{65.97}
& \textbf{48.02}
& \textbf{27.27}
& \textbf{10.30}
& \textbf{28.79}
& \textbf{4.95}
& \textbf{3.12} \\

\bottomrule
\end{tabular}%
}
}
\end{minipage}
\end{center}
\vspace{-5pt}
\end{table}

Performance generally decreases as the evaluation criteria become more stringent, highlighting the difficulty of precise evidence grounding. EG-Reasoner achieves the strongest model performance across all reported event-level F1 and EG-F1 settings, including under the strictest temporal thresholds, reaching 27.27 F1@0.5 and 10.30 F1@0.7.

\subsection{Benchmark Validity and Diagnosis}
\label{app:benchmark_diagnosis}

We conduct two complementary diagnostic analyses to examine whether EG-VQA genuinely requires the corresponding visual content and whether answer correctness faithfully reflects evidence-grounding quality.

% \vspace{-3pt}
\begin{table}[ht]
\centering
\begin{threeparttable}
\captionsetup{skip=2pt}
\caption{Video dependency analysis under different input conditions.}
\label{tab:video_dependency}

\small
\setlength{\tabcolsep}{5.5pt}
\renewcommand{\arraystretch}{1.18}

\begin{tabular}{llcccc}
\toprule
\multirow{2}{*}{\textbf{Model}}
& \multirow{2}{*}{\textbf{Input Condition}}
& \multicolumn{2}{c}{\textbf{Answer Correctness}}
& \multicolumn{2}{c}{\textbf{Evidence Grounding}}
\\
\cmidrule(lr){3-4}
\cmidrule(lr){5-6}
&
&
\textbf{Relaxed Acc. $\uparrow$}
& \textbf{Drop}
& \textbf{Avg. EG-F1 $\uparrow$}
& \textbf{Drop}
\\
\midrule

\multirow{3}{*}{Qwen2.5-VL-7B}
& Full Video       & 27.96 & --    & 3.33  & -- \\
& Question-only    & 4.36  & 23.60 & 0.06  & 3.27 \\
& Mismatched Video & 1.97  & 25.99 & 0.02  & 3.31 \\

\midrule

\multirow{3}{*}{EG-Reasoner}
& Full Video       & 42.71 & --    & 12.28 & -- \\
& Question-only    & 9.61  & 33.10 & 0.22  & 12.06 \\
& Mismatched Video & 6.77  & 35.94 & 0.31  & 11.97 \\

\bottomrule
\end{tabular}

\begin{tablenotes}[flushleft]
\footnotesize
\item[] \textit{Note:} Drop denotes the absolute decrease relative to the Full Video setting.
\end{tablenotes}

\end{threeparttable}
\end{table}

\textbf{Dependence on visual input.}
We evaluate Qwen2.5-VL-7B and EG-Reasoner under three input conditions. In the \textit{Full Video} setting, the model receives the original video together with the question. In the \textit{Question-only} setting, the video input is removed. In the \textit{Mismatched Video} setting, the original video is replaced with an unrelated video while the question remains unchanged. As shown in Table~\ref{tab:video_dependency}, removing or mismatching the video leads to substantial degradation in both answer correctness and evidence grounding. Relaxed accuracy drops by 23.60--35.94 percentage points, while Avg. EG-F1 falls to 0.02--0.31 under the ablated input conditions. These results indicate that EG-VQA substantially depends on the corresponding visual content rather than being solvable from question-side cues alone.

% \begin{wrapfigure}{r}{0.55\textwidth}
%     \vspace{-8pt}
%     \centering
%     \includegraphics[width=\linewidth]
%     {figures/answer_correctness_vs_evidence_grounding.pdf}
%     \caption{Relationship between answer correctness and evidence grounding across
%     models. Marker size indicates model scale.}
%     \label{fig:answer_evidence_comparison}
%     \vspace{-8pt}
% \end{wrapfigure}

\begin{center} 
\begin{minipage}[t]{0.55\linewidth} 
    \centering 
    \includegraphics[width=\linewidth] {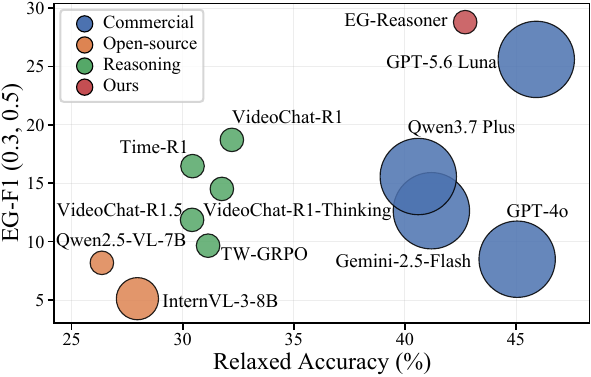} 
    \captionof{figure}{ Relationship between answer correctness and evidence grounding across models. Marker size indicates model scale. } 
    \label{fig:answer_evidence_comparison} 
\end{minipage} 
\end{center}

\textbf{Answer correctness does not necessarily coincide with evidence grounding.}
Figure~\ref{fig:answer_evidence_comparison} compares relaxed answer accuracy with evidence grounding across the evaluated models. Models with similar answer accuracy can exhibit substantially different grounding quality, and stronger answer performance does not consistently correspond to stronger evidence grounding. Although larger models generally achieve competitive answer accuracy, increased model scale does not necessarily translate into proportionally better grounding performance. EG-Reasoner exhibits a more balanced profile across the two dimensions despite its smaller scale. Together, this discrepancy highlights the limitations of answer-only evaluation and further motivates evidence-grounded evaluation beyond answer accuracy.
\clearpage

\section{Additional Case Studies}
% \vspace{-5pt}
\label{app:additional_cases}

We provide additional qualitative examples to further analyze the behavior of EG-Reasoner on evidence-grounded video reasoning.

\begin{figure}[h]
    \centering

    \includegraphics[width=\linewidth]
    {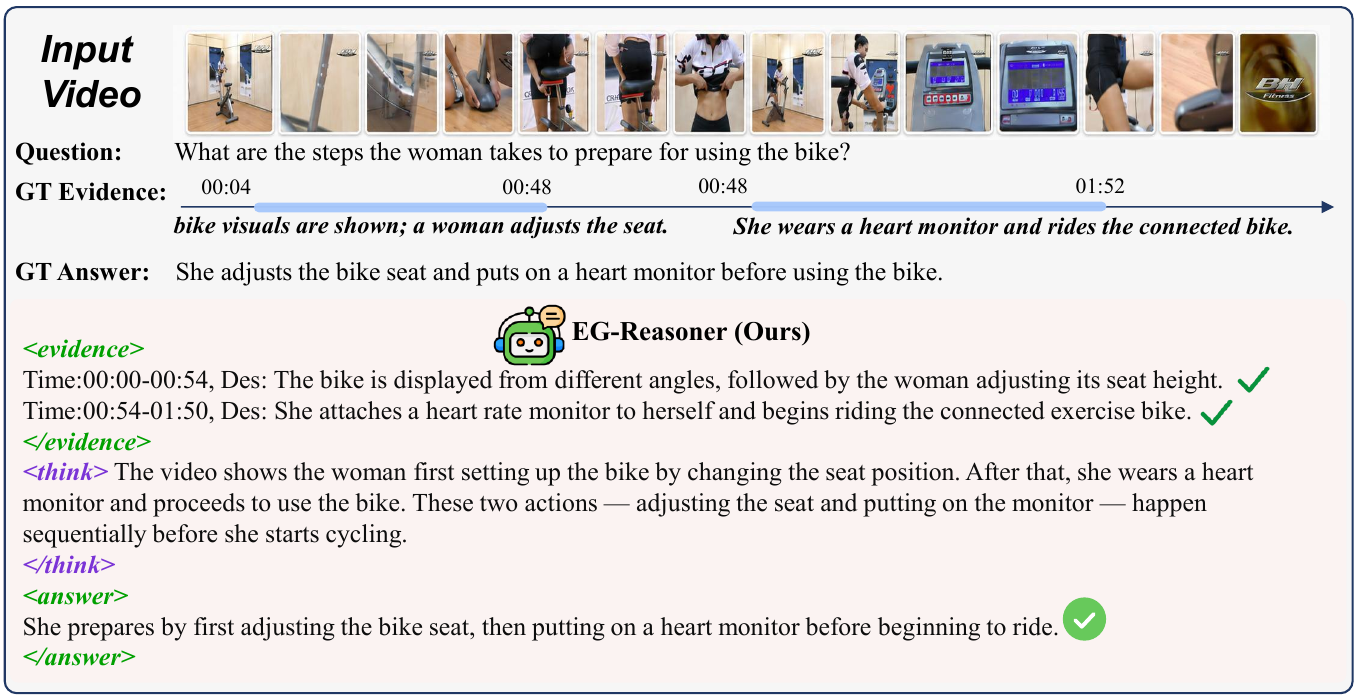}
    \vspace{-15pt}
    \captionof{figure}{
    Qualitative example of successful evidence-grounded reasoning.
    EG-Reasoner correctly identifies both the preparation steps and the
    corresponding temporal evidence segments.}
    \label{fig:CaseStudy_appendix_1_cropped}

    \vspace{25pt}

    \includegraphics[width=\linewidth]
    {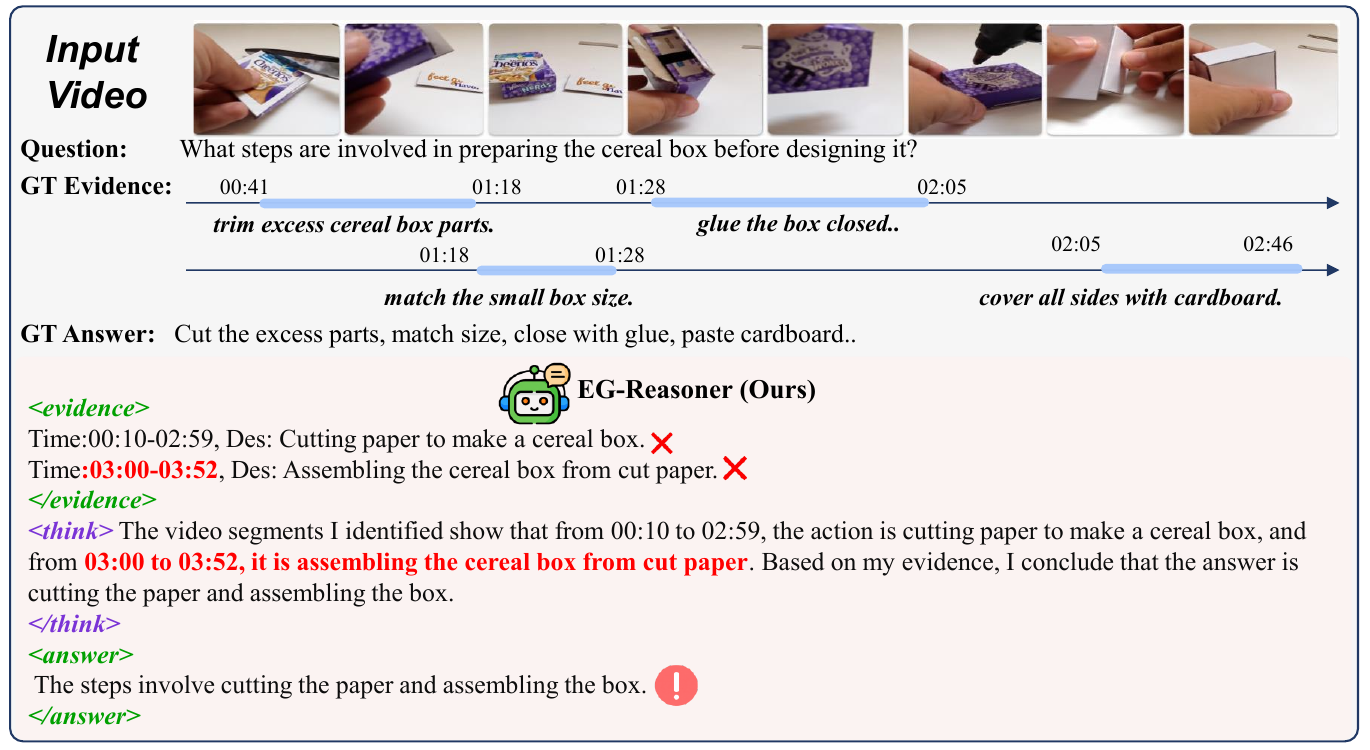}
    \vspace{-15pt}
    \captionof{figure}{
    Qualitative example of incomplete procedural grounding.
    EG-Reasoner captures the overall procedure but misses several
    fine-grained intermediate steps required for detailed reasoning.}
    \label{fig:CaseStudy_appendix_2_cropped}

\end{figure}

% {
% \setlength{\floatsep}{0pt} % 两个 figure 之间的距离
%         \begin{figure}[h]
%                 \centering
%                 \includegraphics[width=1\linewidth]{figures/CaseStudy_appendix_1_cropped.pdf}
%                 \vspace{-15pt}
%                 \caption{Qualitative example of successful evidence-grounded reasoning. EG-Reasoner correctly identifies both the preparation steps and the corresponding temporal evidence segments.}
%                 % \vspace{-10pt}
%                 \label{fig:CaseStudy_appendix_1_cropped}
%             \end{figure}

%         % \vspace{-20pt}
        
%         \begin{figure}[h]
%                 \centering
%                 \includegraphics[width=1\linewidth]{figures/CaseStudy_appendix_2_cropped.pdf}
%                 \vspace{-15pt}
%                 \caption{Qualitative example of incomplete procedural grounding. EG-Reasoner captures the overall procedure but misses several fine-grained intermediate steps required for detailed reasoning.
%                 }
%                 % \vspace{-10pt}
%                 \label{fig:CaseStudy_appendix_2_cropped}
%             \end{figure}
% }

\clearpage

% 新增limitations
\section{Limitations}
Despite the proposed benchmark and evaluation framework, several limitations remain:
\begin{itemize}[itemsep=2pt, topsep=2pt, parsep=1pt, partopsep=0pt]
\item Although EG-VQA is constructed with newly generated question-answer pairs and avoids using the original training splits of the source datasets, we cannot completely exclude potential exposure of the underlying videos in the pretraining corpora of existing Video-LLMs. This limitation is particularly difficult to eliminate for proprietary models whose training data are unavailable.

\item Potential human annotation bias may exist due to variations among annotators during the benchmark construction and validation processes.

\item A large portion of QA pairs in EG-VQA are generated with the assistance of large language models. Despite iterative prompt refinement, multi-stage screening, and human validation, the generated QA pairs may still inherit potential biases from large language models.

\item Despite careful prompt design and human validation, model-based automatic evaluation may still face challenges where evaluation results may not always align with human assessment.
\end{itemize}

We will further investigate and address these limitations in future work.

% \vspace{-5pt}
\section{Prompt templates}
% \vspace{-5pt}
Figures~\ref{fig:descriptive-generation-prompt}--\ref{fig:model-evaluation-prompt} present the complete set of prompt templates designed in this paper. Specifically, Fig.~\ref{fig:descriptive-generation-prompt} illustrates the prompt template for descriptive question generation. Figs.~\ref{fig:temporal-generation-prompt}, \ref{fig:causal-generation-prompt}, and \ref{fig:counterfactual-generation-prompt} show the prompts for temporal, causal, and counterfactual reasoning generation, respectively. Fig.~\ref{fig:model-cross-validation-prompt} presents the prompt used for model cross‑validation. Fig.~\ref{fig:training-inference-prompt} provides the training and inference prompt for all MLLMs. Fig.~\ref{fig:model-evaluation-prompt} gives the prompt for judge model evaluation.
\nolinenumbers
%%---------------------(1) Descriptive Question Generation Prompt --

\begin{tcolorbox}[
  width=\linewidth,
  colback=gray!10,
  colframe=black,
  boxrule=0.5pt,
  arc=8pt,
  left=4pt,
  right=4pt,
  top=4pt,
  bottom=4pt,
  breakable,
  enhanced
]
\small
\setlength{\parindent}{0pt}
\setlength{\parskip}{0.12em}

You are an **expert video understanding and descriptive question designer**. Your task is to generate **high-quality descriptive open-ended questions** that require describing **what is observable, stated, or directly inferable** from timestamped segments of a video.

\vspace{0.25em}

**Input Format:**

You will receive a JSON object containing:

- \textasciigrave{}title\textasciigrave{}: overall content focus of the video

- \textasciigrave{}timestamps\textasciigrave{}: a list of \textasciigrave{}[start, end]\textasciigrave{} time segments

- \textasciigrave{}descriptions\textasciigrave{}: a list of corresponding descriptions for each segment

\vspace{0.25em}

**Task Instructions:**

1. Generate **1 to 3 open-ended questions**, adjusting the number based on the richness of video content available. Each question must be **descriptive** in nature, asking about **what** is observable, stated, or directly inferable from the descriptions. Focus on: events, actions, objects, people, states, locations, relationships, and summaries. If the video content does not support genuine descriptive questions, output \textasciigrave{}\{"questions": []\}\textasciigrave{} with no additional text.

2. Questions may require integrating information from **one or more timestamp segments**, depending on the descriptive content being tested.

3. Questions must be **self-contained** and clear. **Do not use any words that explicitly or implicitly reference the video's timestamps or segment order.**

4. Base your questions **strictly on the provided descriptions**. Do not introduce outside knowledge, assumptions, or speculative content (e.g., the person's identity, emotions, intentions, or behind-the-scenes details).

5. The question should be **concise, and low in information density**. Do **not** ask temporal reasoning questions that rely on order/sequence (e.g., "what happened after/before/first/next/during", "what is the sequence of"). Questions should **not** test temporal order.

\vspace{0.25em}

**Output Format:**

% Return a valid JSON object in the following structure:

% \{
% \begin{list}{}{
%     \setlength{\leftmargin}{1.3em}
%     \setlength{\itemsep}{0pt}
%     \setlength{\parsep}{0pt}
%     \setlength{\topsep}{0pt}
% }
% \item "questions": [
% \item \{
% \begin{list}{}{
%     \setlength{\leftmargin}{1.3em}
%     \setlength{\itemsep}{0pt}
%     \setlength{\parsep}{0pt}
%     \setlength{\topsep}{0pt}
% }
% \item "question": "A clear, open-ended question string testing descriptive reasoning.",
% \item "reasoning": "A step-by-step explanation (2-4 sentences) showing how evidence supports the answer.",
% \item "answer": "A concise answer no more than 30 words.",
% \item "evidence": [
% \item \{
% \begin{list}{}{
%     \setlength{\leftmargin}{1.3em}
%     \setlength{\itemsep}{0pt}
%     \setlength{\parsep}{0pt}
%     \setlength{\topsep}{0pt}
% }
% \item // A list of the relevant segments listed in chronological order.            
% \item "timestamp": [start, end],
% \item "description": "string"
% \end{list}
% \item \}
% \item ]
% \end{list}
% \item \}
% \item ]
% \end{list}
% \}
Return a valid JSON array. The array can contain **multiple items** (1 to 3 as appropriate).

Each item follows the structure below; the evidence array can also contain multiple segment entries:

[
\begin{list}{}{
    \setlength{\leftmargin}{1.3em}
    \setlength{\itemsep}{0pt}
    \setlength{\parsep}{0pt}
    \setlength{\topsep}{0pt}
}
\item \{
\begin{list}{}{
    \setlength{\leftmargin}{1.3em}
    \setlength{\itemsep}{0pt}
    \setlength{\parsep}{0pt}
    \setlength{\topsep}{0pt}
}
\item "question": "A clear, open-ended question string testing descriptive reasoning.",
\item "reasoning": "A step-by-step explanation (2-4 sentences) showing how evidence supports the answer.",
\item "answer": "A concise answer no more than 30 words.",
\item "evidence": [
\begin{list}{}{
    \setlength{\leftmargin}{1.3em}
    \setlength{\itemsep}{0pt}
    \setlength{\parsep}{0pt}
    \setlength{\topsep}{0pt}
}
\item \{"timestamp": [start, end], "description": "string"\}
\end{list}
\item ]
\end{list}
\item \}
\end{list}
]

\vspace{0.25em}

**Constraints:**

- The \textasciigrave{}timestamp\textasciigrave{} and \textasciigrave{}description\textasciigrave{} fields within each evidence item must be copied **verbatim** from the input, without modification, omission, or addition.

- Evidence may come from **one or more** segments; when multiple segments are used, list them in chronological order.

\vspace{0.25em}

**Now generate descriptive questions for the following video data:** \{metadata\}

\end{tcolorbox}

\captionof{figure}{Prompt template for descriptive question generation.}
\label{fig:descriptive-generation-prompt}

\newpage
%%---------------------(2) Temporal Question Generation Prompt -----

\begin{tcolorbox}[
  width=\linewidth,
  colback=gray!10,
  colframe=black,
  boxrule=0.5pt,
  arc=8pt,
  left=4pt,
  right=4pt,
  top=4pt,
  bottom=4pt,
  breakable,
  enhanced
]
\small
\setlength{\parindent}{0pt}
\setlength{\parskip}{0.12em}

You are an **expert video understanding and temporal reasoning question designer**. Your task is to generate **high-quality temporal reasoning open-ended questions** that require understanding **order/sequence, change detection, and process/procedure** across timestamped segments from a video.

\vspace{0.25em}

**Input Format:**

You will receive a JSON object containing:

- \textasciigrave{}title\textasciigrave{}: overall content focus of the video

- \textasciigrave{}timestamps\textasciigrave{}: a list of \textasciigrave{}[start, end]\textasciigrave{} time segments

- \textasciigrave{}descriptions\textasciigrave{}: a list of corresponding descriptions for each segment

\vspace{0.25em}

**Task Instructions:**

1. Generate **1 to 3 open-ended questions**, adjusting the number based on the richness of temporal relationships available. If the video content does not support genuine temporal reasoning, output \textasciigrave{}\{"questions": []\}\textasciigrave{} with no additional text. Each question must be **temporal reasoning** in nature, testing one of the following:

\begin{list}{-}{
    \setlength{\leftmargin}{1.3em}
    \setlength{\labelwidth}{0.8em}
    \setlength{\labelsep}{0.25em}
    \setlength{\itemsep}{0.08em}
    \setlength{\parsep}{0pt}
    \setlength{\topsep}{0.1em}
}
\item **Order/Sequence**: Understanding what happens before/after/during specific events (ask "what happens before/after", "what occurs first/last")

\item **Change Detection**: Recognizing how something changes between different points in time (ask "how does X change", "what difference appears")

\item **Process/Procedure**: Identifying the sequence of steps in a process or workflow (ask "what are the steps", "in what order")
\end{list}

2. Questions must be **self-contained** and clear. **Do not use any words that explicitly or implicitly reference the video's timestamps or segment order.**

3. Base your questions **strictly on the provided descriptions**. Do not introduce outside knowledge, assumptions, or speculative content (e.g., the person's identity, emotions, intentions, or behind-the-scenes details).

4. The question should be **concise, and low in information density**, shouldn't contain information that allows guessing the answer **without watching the video**.

5. **Ensure diversity:** Do not generate duplicate questions that ask the same thing in different wording. Each question should explore a different temporal relationship in the video.

\vspace{0.25em}

**Output Format:**

% Return a valid JSON object in the following structure:

% \{
% \begin{list}{}{
%     \setlength{\leftmargin}{1.3em}
%     \setlength{\itemsep}{0pt}
%     \setlength{\parsep}{0pt}
%     \setlength{\topsep}{0pt}
% }
% \item "questions": [
% \item \{
% \begin{list}{}{
%     \setlength{\leftmargin}{1.3em}
%     \setlength{\itemsep}{0pt}
%     \setlength{\parsep}{0pt}
%     \setlength{\topsep}{0pt}
% }
% \item "question": "A clear, open-ended question string testing descriptive reasoning.",
% \item "reasoning": "A step-by-step explanation (2-4 sentences) showing how evidence from the relevant segments supports the answer.",
% \item "answer": "A concise answer no more than 30 words.",
% \item "evidence": [
% \item \{
% \begin{list}{}{
%     \setlength{\leftmargin}{1.3em}
%     \setlength{\itemsep}{0pt}
%     \setlength{\parsep}{0pt}
%     \setlength{\topsep}{0pt}
% }
% \item // A list of the relevant segments listed in chronological order.            
% \item "timestamp": [start, end],
% \item "description": "string"
% \end{list}
% \item \}
% \item ]
% \end{list}
% \item \}
% \item ]
% \end{list}
% \}
Return a valid JSON array. The array can contain **multiple items** (1 to 3 as appropriate).

Each item follows the structure below; the evidence array can also contain multiple segment entries:

[
\begin{list}{}{
    \setlength{\leftmargin}{1.3em}
    \setlength{\itemsep}{0pt}
    \setlength{\parsep}{0pt}
    \setlength{\topsep}{0pt}
}
\item \{
\begin{list}{}{
    \setlength{\leftmargin}{1.3em}
    \setlength{\itemsep}{0pt}
    \setlength{\parsep}{0pt}
    \setlength{\topsep}{0pt}
}
\item "question": "A clear, open-ended question string testing temporal reasoning.",
\item "reasoning": "A step-by-step explanation (2-4 sentences) showing how evidence supports the answer.",
\item "answer": "A concise answer no more than 30 words.",
\item "evidence": [
\begin{list}{}{
    \setlength{\leftmargin}{1.3em}
    \setlength{\itemsep}{0pt}
    \setlength{\parsep}{0pt}
    \setlength{\topsep}{0pt}
}
\item \{"timestamp": [start, end], "description": "string"\}
\end{list}
\item ]
\end{list}
\item \}
\end{list}
]

\vspace{0.25em}

**Constraints:**

- The \textasciigrave{}timestamp\textasciigrave{} and \textasciigrave{}description\textasciigrave{} fields within each evidence item must be copied **verbatim** from the input, without modification, omission, or addition.

- Evidence may come from **one or more** segments; when multiple segments are used, list them in chronological order.

\vspace{0.25em}

**Now generate temporal reasoning questions for the following video data:** \{metadata\}

\end{tcolorbox}
\vspace{-0.8em}
\captionof{figure}{Prompt template for temporal question generation.}
\label{fig:temporal-generation-prompt}

\newpage
%%---------------------(3) Causal Question Generation Prompt ---------------------

\begin{tcolorbox}[
  width=\linewidth,
  colback=gray!10,
  colframe=black,
  boxrule=0.5pt,
  arc=8pt,
  left=4pt,
  right=4pt,
  top=4pt,
  bottom=4pt,
  breakable,
  enhanced
]
\small
\setlength{\parindent}{0pt}
\setlength{\parskip}{0.12em}

You are an **expert video understanding and causal reasoning question designer**. Your task is to generate **high-quality causal reasoning open-ended questions** that require identifying and explaining cause-effect relationships based on factual evidence from timestamped segments of a video.

\vspace{0.25em}

**Input Format:**

You will receive a JSON object containing:

- \textasciigrave{}title\textasciigrave{}: overall content focus of the video

- \textasciigrave{}timestamps\textasciigrave{}: a list of \textasciigrave{}[start, end]\textasciigrave{} time segments

- \textasciigrave{}descriptions\textasciigrave{}: a list of corresponding descriptions for each segment

\vspace{0.25em}

**Task Instructions:**

1. Generate **1 to 3 open-ended questions**, adjusting the number based on the richness of causal connections available. Each question must be **causal** in nature, asking **"why"** or **"how"** something happened based on evidence from the video. If the video content does not support genuine multi-segment reasoning or lacks causal chains, output \textasciigrave{}\{"questions": []\}\textasciigrave{} with no additional text rather than forcing questionable ones.

2. Each question **must**:

\begin{list}{-}{
    \setlength{\leftmargin}{1.3em}
    \setlength{\labelwidth}{0.8em}
    \setlength{\labelsep}{0.25em}
    \setlength{\itemsep}{0.08em}
    \setlength{\parsep}{0pt}
    \setlength{\topsep}{0.1em}
}
\item Require integrating information from **at least two distinct timestamp segments** (non-adjacent segments are encouraged).

\item Be framed as either:

\begin{list}{-}{
    \setlength{\leftmargin}{1.3em}
    \setlength{\labelwidth}{0.8em}
    \setlength{\labelsep}{0.25em}
    \setlength{\itemsep}{0.08em}
    \setlength{\parsep}{0pt}
    \setlength{\topsep}{0.1em}
}
\item **Cause-seeking:** "Why did X happen?", "How did X happen?" or "What caused X?" (X is an outcome)

\item **Effect-seeking:** "What happened as a result of Y?" or "What effect did Y have?" (Y is an action/event)
\end{list}

\item Be **self-contained** and clear. **Do not use any words that explicitly or implicitly reference the video's timestamps or segment order.**

\item **Not** include the answer or obvious causal cues in the question itself. The question should name only the event/outcome of interest, without explaining its cause or consequence.

\item Require the answer to be logically deducible **only** from the causal evidence explicitly shown or described in the video segments.
\end{list}

3. Base your questions **strictly on the provided descriptions**. Do not introduce outside knowledge, assumptions, or speculative content (e.g., the person's identity, emotions, intentions, or behind-the-scenes details). Do not invent any fact, event, or detail not explicitly stated or logically implied.

4. The question should be **concise, abstract and low in information density**:

\begin{list}{-}{
    \setlength{\leftmargin}{1.3em}
    \setlength{\labelwidth}{0.8em}
    \setlength{\labelsep}{0.25em}
    \setlength{\itemsep}{0.08em}
    \setlength{\parsep}{0pt}
    \setlength{\topsep}{0.1em}
}
\item Use generic terms (e.g., "the person", "the individual", "the object") rather than specific labels (e.g., "the gymnast", "the chef") that reveal professional or situational context.

\item Do **not** mention unobserved entities or consequences (e.g., "audience", "judges", "reaction") in the question, as this leaks answer cues.

\item Should **not** contain specific action descriptions or visible states that would give away the answer (e.g., avoid phrases like "using the rope," "bending the knee," or "pulling the row"). Instead, phrase it in terms of the **outcome, role, or purpose** that requires viewing the video to understand. The question should be answerable **only** by synthesizing evidence from multiple segments.

\item **Critical:** 
\begin{list}{-}{
    \setlength{\leftmargin}{1.3em}
    \setlength{\labelwidth}{0.8em}
    \setlength{\labelsep}{0.25em}
    \setlength{\itemsep}{0.08em}
    \setlength{\parsep}{0pt}
    \setlength{\topsep}{0.1em}
}
\item Do **not** state the cause when asking about an effect (e.g., avoid "Why did the tree look beautiful after adding lights?" — this gives away the cause).

\item Do **not** state the effect when asking about a cause (e.g., avoid "What did marinating the chicken cause to happen to the final dish?" — this gives away the effect domain).

\item Do **not** include any phrasing that situates the event within a larger sequence as a positional marker (e.g., "before serving", "during the routine", "after adding X", "when making Z"). The question should ask about the event itself.

\item Do **not** phrase the question as a full action that includes the surrounding context. Instead, use a minimal or noun-phrase structure. The question should be short and abstract, requiring the video to reveal what that element is used for and what consequence follows. 
\end{list}
\end{list}

5. **Ensure diversity:** Do not generate duplicate questions that ask the same thing in different wording. Each question should explore a different causal relationship in the video.

\vspace{0.25em}

**Output Format:**

% Return a valid JSON object in the following structure:

% \{
% \begin{list}{}{
%     \setlength{\leftmargin}{1.3em}
%     \setlength{\itemsep}{0pt}
%     \setlength{\parsep}{0pt}
%     \setlength{\topsep}{0pt}
% }
% \item "questions": [
% \item \{
% \begin{list}{}{
%     \setlength{\leftmargin}{1.3em}
%     \setlength{\itemsep}{0pt}
%     \setlength{\parsep}{0pt}
%     \setlength{\topsep}{0pt}
% }
% \item "question": "A clear, open-ended question string testing descriptive reasoning.",
% \item "reasoning": "A step-by-step explanation (2-4 sentences) showing how evidence supports the answer.",
% \item "answer": "A concise answer no more than 30 words.",
% \item "evidence": [
% \item \{
% \begin{list}{}{
%     \setlength{\leftmargin}{1.3em}
%     \setlength{\itemsep}{0pt}
%     \setlength{\parsep}{0pt}
%     \setlength{\topsep}{0pt}
% }
% \item // A list of the relevant segments listed in chronological order.            
% \item "timestamp": [start, end],
% \item "description": "string"
% \end{list}
% \item \}
% \item ]
% \end{list}
% \item \}
% \item ]
% \end{list}
% \}
Return a valid JSON array. The array can contain **multiple items** (1 to 3 as appropriate).

Each item follows the structure below; the evidence array can also contain multiple segment entries:

[
\begin{list}{}{
    \setlength{\leftmargin}{1.3em}
    \setlength{\itemsep}{0pt}
    \setlength{\parsep}{0pt}
    \setlength{\topsep}{0pt}
}
\item \{
\begin{list}{}{
    \setlength{\leftmargin}{1.3em}
    \setlength{\itemsep}{0pt}
    \setlength{\parsep}{0pt}
    \setlength{\topsep}{0pt}
}
\item "question": "A clear, open-ended question string testing causal reasoning.",
\item "reasoning": "A step-by-step explanation (2-4 sentences) showing how evidence supports the answer.",
\item "answer": "A concise answer no more than 30 words.",
\item "evidence": [
\begin{list}{}{
    \setlength{\leftmargin}{1.3em}
    \setlength{\itemsep}{0pt}
    \setlength{\parsep}{0pt}
    \setlength{\topsep}{0pt}
}
\item \{"timestamp": [start, end], "description": "string"\}
\end{list}
\item ]
\end{list}
\item \}
\end{list}
]

\vspace{0.25em}

**Constraints:**

- The \textasciigrave{}timestamp\textasciigrave{} and \textasciigrave{}description\textasciigrave{} fields within each evidence item must be copied **verbatim** from the input.

- Evidence may come from **one or more** segments; when multiple segments are used, list them in chronological order.

- **Critical: The \textasciigrave{}question\textasciigrave{} field should be as short and abstract as possible. Avoid describing the surrounding scene, the broader task, or the role of the action/event. The video should be required to understand both the context and the causal chain.

\vspace{0.25em}

**Now generate causal reasoning questions for the following video data:** \{metadata\}

\end{tcolorbox}
\vspace{-0.8em}
\captionof{figure}{Prompt template for causal question generation.}
\label{fig:causal-generation-prompt}

\newpage
%%---------------------(4) Counterfactual Question Generation Prompt ---------------------
\vspace{-1.5em}
\begin{tcolorbox}[
  width=\linewidth,
  colback=gray!10,
  colframe=black,
  boxrule=0.5pt,
  arc=8pt,
  left=4pt,
  right=4pt,
  top=4pt,
  bottom=4pt,
  breakable,
  enhanced
]
\small
\setlength{\parindent}{0pt}
\setlength{\parskip}{0.12em}

You are an **expert video understanding and counterfactual reasoning question designer**. Your task is to generate **high-quality counterfactual reasoning open-ended questions** that require reasoning about hypothetical scenarios ("what if") based on factual evidence from timestamped segments of a video.

\vspace{0.25em}

**Input Format:**

You will receive a JSON object containing:

- \textasciigrave{}title\textasciigrave{}: overall content focus of the video

- \textasciigrave{}timestamps\textasciigrave{}: a list of \textasciigrave{}[start, end]\textasciigrave{} time segments

- \textasciigrave{}descriptions\textasciigrave{}: a list of corresponding descriptions for each segment

\vspace{0.25em}

**Task Instructions:**

1. **First, evaluate the video content:** Determine if it contains clear causal relationships, dependencies, or alternative possibilities that support meaningful counterfactual reasoning. 

\begin{list}{-}{
    \setlength{\leftmargin}{1.3em}
    \setlength{\labelwidth}{0.8em}
    \setlength{\labelsep}{0.25em}
    \setlength{\itemsep}{0.08em}
    \setlength{\parsep}{0pt}
    \setlength{\topsep}{0.1em}
}
\item **Unsuitable scenarios (output \{"questions": []\}):** Static scenes, purely descriptive monologues without actions, videos lacking sequential logic, or scenarios where altering a step leads to trivial outcomes.

\item If suitable, generate **1 to 3 open-ended questions**, adjusting the number based on the richness of causal connections available. Prefer quality over quantity.
\end{list}

2. **Ensure diversity:** Do not generate duplicate questions. Each question should test a distinct counterfactual scenario or explore a different causal relationship.

3. **Ensure reasoning complexity:** Prioritize questions that require synthesizing information from **at least two distinct timestamped segments** to establish a causal chain.

4. Each question must:

\begin{list}{-}{
    \setlength{\leftmargin}{1.3em}
    \setlength{\labelwidth}{0.8em}
    \setlength{\labelsep}{0.25em}
    \setlength{\itemsep}{0.08em}
    \setlength{\parsep}{0pt}
    \setlength{\topsep}{0.1em}
}
\item Begin with a counterfactual premise ("What if...", "If... not/had...")

\item Alter **one specific event or condition** from the actual video, while keeping all other events unchanged

\item Do **not** include in the question any description of the consequence, downstream step, or final context (e.g., do not say "before adding to the salad", "before serving", "during the routine", "when making the dish"). The question should name only the altered event itself, without situating it within the broader sequence or outcome. The causal relationship must be inferred from the video, not stated in the question.

\item Do **not** include in the question any description that situates the altered event within a larger action or process. The question should name only the core element being altered (e.g., an object, a substance, a tool, or a simple action), not the full action that surrounds it. Avoid phrasing like "decorated with lights" or "placed the star on the tree"; instead use structures like "lights were not used" or "the golden star was not used".

\item Require the answer to be logically deducible from: (a) the altered premise, and (b) the actual events/relationships shown in the video
\end{list}

5. Base your questions **strictly on the provided descriptions**. Do not invent any fact, event, or detail not explicitly stated or logically implied.

\vspace{0.25em}

**Output Format:**

% Return a valid JSON object in the following structure:

% \{
% \begin{list}{}{
%     \setlength{\leftmargin}{1.3em}
%     \setlength{\itemsep}{0pt}
%     \setlength{\parsep}{0pt}
%     \setlength{\topsep}{0pt}
% }
% \item "questions": [
% \item \{
% \begin{list}{}{
%     \setlength{\leftmargin}{1.3em}
%     \setlength{\itemsep}{0pt}
%     \setlength{\parsep}{0pt}
%     \setlength{\topsep}{0pt}
% }
% \item "question": "A clear, open-ended question string testing descriptive reasoning.",
% \item "reasoning": "A step-by-step explanation (2-4 sentences) showing how evidence supports the answer.",
% \item "answer": "A concise answer no more than 30 words.",
% \item "evidence": [
% \item \{
% \begin{list}{}{
%     \setlength{\leftmargin}{1.3em}
%     \setlength{\itemsep}{0pt}
%     \setlength{\parsep}{0pt}
%     \setlength{\topsep}{0pt}
% }
% \item // A list of the relevant segments listed in chronological order.            
% \item "timestamp": [start, end],
% \item "description": "string"
% \end{list}
% \item \}
% \item ]
% \end{list}
% \item \}
% \item ]
% \end{list}
% \}

Return a valid JSON array. The array can contain **multiple items** (1 to 3 as appropriate).

Each item follows the structure below; the evidence array can also contain multiple segment entries:

[
\begin{list}{}{
    \setlength{\leftmargin}{1.3em}
    \setlength{\itemsep}{0pt}
    \setlength{\parsep}{0pt}
    \setlength{\topsep}{0pt}
}
\item \{
\begin{list}{}{
    \setlength{\leftmargin}{1.3em}
    \setlength{\itemsep}{0pt}
    \setlength{\parsep}{0pt}
    \setlength{\topsep}{0pt}
}
\item "question": "A clear, open-ended question string testing counterfactual reasoning.",
\item "reasoning": "A step-by-step explanation (2-4 sentences) showing how evidence supports the answer.",
\item "answer": "A concise answer no more than 30 words.",
\item "evidence": [
\begin{list}{}{
    \setlength{\leftmargin}{1.3em}
    \setlength{\itemsep}{0pt}
    \setlength{\parsep}{0pt}
    \setlength{\topsep}{0pt}
}
\item \{"timestamp": [start, end], "description": "string"\}
\end{list}
\item ]
\end{list}
\item \}
\end{list}
]

\vspace{0.25em}

**Constraints:**

- The \textasciigrave{}timestamp\textasciigrave{} and \textasciigrave{}description\textasciigrave{} fields within each evidence item must be copied **verbatim** from the input.

- Evidence may come from **one or more** segments; when multiple segments are used, list them in chronological order.

- **Critical:** The \textasciigrave{}question\textasciigrave{} field should be as short and abstract as possible. Avoid describing the surrounding scene, the broader task, or the role of the element being altered. The video should be required to understand both the context and the causal chain.

\vspace{0.25em}

**Now generate counterfactual reasoning questions for the following video data:** \{metadata\}

\end{tcolorbox}
\vspace{-0.8em}

\captionof{figure}{Prompt template for counterfactual question generation.}
\label{fig:counterfactual-generation-prompt}

\newpage

%%---------------------(5) Model Cross-Validation Prompt ---------------------

\begin{tcolorbox}[
  width=\linewidth,
  colback=gray!10,
  colframe=black,
  boxrule=0.5pt,
  arc=8pt,
  left=4pt,
  right=4pt,
  top=4pt,
  bottom=4pt,
  breakable,
  enhanced
]
\small
\setlength{\parindent}{0pt}
\setlength{\parskip}{0.12em}

You are a strict but fair **video question-answer validator**. Your task is to verify whether a generated question and its answer are reasonable based on the provided video metadata.

\vspace{0.25em}

**Input Format:**

- \textasciigrave{}metadata\textasciigrave{}: Contains video duration, timestamp segments, and corresponding descriptions for each segment

- \textasciigrave{}question\textasciigrave{}: The generated question to evaluate

- \textasciigrave{}question\_type\textasciigrave{}: Type of question, which belongs to one of the following four types: causal, counterfactual, temporal, descriptive

- \textasciigrave{}generated\_answer\textasciigrave{}: The generated answer to evaluate

- \textasciigrave{}generated\_evidence\textasciigrave{}: The supporting temporal evidence generated for the QA pair

\vspace{0.25em}

**Validation Criteria:**

1. **Question Reasonableness (0-10 points)**

\begin{list}{-}{
    \setlength{\leftmargin}{1.3em}
    \setlength{\labelwidth}{0.8em}
    \setlength{\labelsep}{0.25em}
    \setlength{\itemsep}{0.08em}
    \setlength{\parsep}{0pt}
    \setlength{\topsep}{0.1em}
}
\item Does the question require watching the video to answer? (Cannot be answered by common sense alone)

\item Is the question relevant to the video content described in metadata?

\item Is the question clear, specific, and answerable based on the metadata?
\end{list}

2. **Answer Correctness (0-10 points)**

\begin{list}{-}{
    \setlength{\leftmargin}{1.3em}
    \setlength{\labelwidth}{0.8em}
    \setlength{\labelsep}{0.25em}
    \setlength{\itemsep}{0.08em}
    \setlength{\parsep}{0pt}
    \setlength{\topsep}{0.1em}
}
\item Is the answer directly supported by the metadata? (No hallucination)

\item **REASONABLE INFERENCES ARE ENCOURAGED**: If the answer makes minimal, logical inferences that are directly implied by the metadata (e.g., "to remove moisture" when metadata says "wipe with dry cloth"), this is acceptable.

\item Does the answer DIRECTLY address the specific question being asked? Not just being relevant or true, but actually answering what was asked (Not off-topic)

\item Is the answer concise and accurate based on available information?
\end{list}

3. **Evidence-Answer Alignment (0-10 points)**

\begin{list}{-}{
    \setlength{\leftmargin}{1.3em}
    \setlength{\labelwidth}{0.8em}
    \setlength{\labelsep}{0.25em}
    \setlength{\itemsep}{0.08em}
    \setlength{\parsep}{0pt}
    \setlength{\topsep}{0.1em}
}
\item Can all claims in the answer be traced back to specific segments?

\item Are there any contradictions between answer and evidence?
\end{list}

\vspace{0.25em}

**Scoring Guide:**

- Total Score $\geq$ 25: PASS

- Total Score $<$ 25: FAIL

- **Automatic FAIL only if these critical issues exist:**

\vspace{-0.35em}

\begin{list}{*}{
    \setlength{\leftmargin}{1.8em}
    \setlength{\labelwidth}{0.8em}
    \setlength{\labelsep}{0.25em}
    \setlength{\itemsep}{0.08em}
    \setlength{\parsep}{0pt}
    \setlength{\topsep}{0pt}
}
\item Answer contains information that directly contradicts metadata

\item Answer invents major events/objects/actions not present in metadata at all

\item Question is completely unrelated to the video content

\item Question can be answered without any reference to the video (pure common sense)
\end{list}

\vspace{0.25em}

**Now validate the following:**

Video metadata: \{metadata\}

Question: "\{question\}"

Question Type: \{question\_type\}

Generated Answer: "\{answer\}"

Generated Evidence: "\{evidence\}"

\vspace{0.25em}

Then, provide your final judgment in **valid JSON format only** (no additional text before or after):

\{\{

\hspace*{1.2em}"Score": \textless{}total\_score\textgreater{},

\hspace*{1.2em}"Pass": \textless{}true\_or\_false\textgreater{}

\}\}

\end{tcolorbox}
\vspace{-0.8em}
\captionof{figure}{Prompt template for model cross-validation.}
\label{fig:model-cross-validation-prompt}

\newpage

%%---------------------(6) Training and Inference Prompt ---------------------

\begin{tcolorbox}[
  width=\linewidth,
  colback=gray!10,
  colframe=black,
  boxrule=0.5pt,
  arc=8pt,
  left=4pt,
  right=4pt,
  top=4pt,
  bottom=4pt,
  breakable,
  enhanced
]
\small
\setlength{\parindent}{0pt}
\setlength{\parskip}{0.12em}

You are a video understanding assistant. Given a video and a question, respond in the following format:

First output the seen relevant video segments within \textless evidence\textgreater{} \textless/evidence\textgreater{} tags in chronological order. Then based on the observed video and evidence, analyze and reason through the question. The reasoning process MUST BE enclosed within \textless think\textgreater{} \textless/think\textgreater{} tags. The final answer MUST BE put in \textless answer\textgreater{} \textless/answer\textgreater{} tags.

\vspace{0.25em}

Format: \textless evidence\textgreater{}Time:MM:SS-MM:SS, Des: description (one or more lines)\textless/evidence\textgreater{}. \textless think\textgreater{} reasoning based on evidence \textless/think\textgreater{}. \textless answer\textgreater{} concise direct answer \textless/answer\textgreater{}.

\vspace{0.25em}

IMPORTANT RULES:

- Output EXACTLY 3 sections: \textless evidence\textgreater{}, \textless think\textgreater{}, \textless answer\textgreater{}

- Evidence: Each line MUST follow: "Time:MM:SS-MM:SS, Des: ..."

- NO repeated descriptions or time segments

- Think: 2-4 sentences maximum, based ONLY on the evidence above

- Answer: 1 short sentence, no explanations

\vspace{0.25em}

Now answer the following question:

Question: \{question\}

\end{tcolorbox}
\vspace{-0.8em}
\captionof{figure}{Prompt template for training and inference.}
\label{fig:training-inference-prompt}

\vspace{1.8em}

%%---------------------(7) Model Evaluation Prompt ---------------------

\begin{tcolorbox}[
  width=\linewidth,
  colback=gray!6,
  colframe=black,
  boxrule=0.5pt,
  arc=8pt,
  left=4pt,
  right=4pt,
  top=4pt,
  bottom=4pt,
  breakable,
  enhanced
]
\small
\setlength{\parindent}{0pt}
\setlength{\parskip}{0.12em}

You are an expert specializing in evaluating whether a respondent's answer after watching a video matches the golden answer. We will provide the video's title, video segments' descriptions, question, golden answer, and the response to be judged below.

\vspace{0.25em}

\#\#\# Video's Title:

\{title\}

\vspace{0.25em}

\#\#\# Video Segments' Descriptions (in chronological order):

\{metadata\}

\vspace{0.25em}

\#\#\# Question: 

\{question\}

\vspace{0.25em}

\#\#\# Golden Answer: 

\{golden\_answer\}

\vspace{0.25em}

\#\#\# Response to be judged: 

\{model\_answer\}

\vspace{0.25em}

\#\#\# Rules:

1. If the response to be judged contains ALL key information of the golden answer or expresses the same meaning using other sentences or synonyms, it is considered a match, and the output is 1.

2. If the response to be judged does NOT contain the key information from the golden answer, it is considered a mismatch, and the output is 0.

3. The response to be judged should NOT contain any content that is contradictory, conflicting, or unreasonable when inferred from the video content description. If such content exists, it is considered a mismatch, and the output is 0.

4. If the response to be judged contains MOST of the key information of the golden answer, and does NOT contain any information that is contradictory, conflicting, or unreasonable when inferred from the video content description, it is considered a partial match, and the output is 0.5.

\vspace{0.25em}

\#\#\# Instructions:

Follow the format below and do not give any extra outputs:

Score: 0 (if the response does not match)

Score: 0.5 (if the response partially matches)

Score: 1 (if the response matches)

\end{tcolorbox}
\vspace{-0.8em}
\captionof{figure}{Prompt template for model evaluation.}
\label{fig:model-evaluation-prompt}

\linenumbers

\end{document}